%% file: main_corl.tex
\documentclass{article}

\usepackage[preprint]{corl_2026}                       

\usepackage{graphicx}
\usepackage{amsmath,amssymb}
\usepackage{booktabs}
\usepackage{multirow}
\usepackage{xcolor}
\usepackage{xspace}
\usepackage{listings}
\usepackage{capt-of}
\usepackage{wrapfig} 
\usepackage{placeins}

\usepackage[disable,colorinlistoftodos,textsize=footnotesize,textwidth=2.2cm,shadow]{todonotes}
\newcommand{\rkk}[1]{\todo[color=yellow!70,size=\tiny,inline=false]{\textbf{rkk:} #1}}
\newcommand{\rkkin}[1]{\todo[color=yellow!40,inline]{\textbf{rkk:} #1}}

\usepackage[ruled,vlined]{algorithm2e}
\DontPrintSemicolon
\SetKwInput{KwIn}{Input}
\SetKwInput{KwOut}{Output}
\SetKwComment{Comment}{$\triangleright$\ }{}

\lstdefinestyle{softflow}{
  basicstyle=\ttfamily\small,
  columns=fullflexible,
  frame=single,
  rulecolor=\color{black!30},
  keywordstyle=\bfseries,
  commentstyle=\itshape\color{black!60},
  showstringspaces=false,
  breaklines=true,
  aboveskip=0.7em,
  belowskip=0.7em
}

\newcommand{\algoname}{SERNF\xspace}

\title{\algoname: Sample-Efficient Real-World Dexterous Fine-Tuning with Normalizing-Flow Policies}

\author{
  Chenyu Yang$^{\ast}$ \\
  Soft Robotics Lab, D-MAVT \\
  ETH Zurich, Switzerland \\
  \texttt{chenyu.yang@srl.ethz.ch} \\
  \And
  Denis Tarasov$^{\ast}$ \\
  Soft Robotics Lab, D-MAVT \\
  ETH Zurich, Switzerland \\
  \texttt{denis.tarasov@srl.ethz.ch} \\
  \And
  Davide Liconti \\
  Soft Robotics Lab, D-MAVT \\
  ETH Zurich, Switzerland \\
  \And
  Romain Guntz \\
  Soft Robotics Lab, D-MAVT \\
  ETH Zurich, Switzerland \\
  \And
  Hehui Zheng \\
  Soft Robotics Lab, D-MAVT \\
  ETH Zurich, Switzerland \\
  \And
  Robert K.~Katzschmann \\
  Soft Robotics Lab, D-MAVT \\
  ETH Zurich, Switzerland \\
}

\begin{document}
\maketitle

\vspace{-2.0\baselineskip}
\begin{abstract}
Real-world fine-tuning of dexterous manipulation policies remains challenging due to limited real-world interaction budgets and highly multimodal action distributions. Diffusion- and flow-matching-based policies, while expressive, do not permit conservative likelihood-based updates during fine-tuning because action probabilities are intractable, and their multi-step denoising makes end-to-end critic-guided fine-tuning costly and unstable on real robots. In contrast, conventional Gaussian policies collapse under multimodality, particularly when actions are executed in chunks, and standard per-step critics fail to align with chunked execution, leading to poor credit assignment. We present \algoname, a sample-efficient off-policy fine-tuning framework with normalizing flow (NF) to address these challenges. The normalizing-flow policy yields exact likelihoods for multimodal action chunks, allowing conservative, stable policy updates through likelihood regularization and thereby improving sample efficiency. An action-chunked critic evaluates entire action sequences, aligning value estimation with the policy's temporal structure and improving long-horizon credit assignment. To our knowledge, this is the first demonstration of a likelihood-based, multimodal generative policy combined with chunk-level value learning on real robotic hardware. We evaluate \algoname on three real-world manipulation tasks: cutting tape with scissors retrieved from a case, in-hand cube rotation with a palm-down grasp, and a rubber-duck pick-and-place -- which together cover precise long-horizon contact, sim-to-real dexterous control, and short-horizon multi-pose grasping with extremely limited data. On these tasks, \algoname achieves stable, sample-efficient adaptation where standard methods struggle.
\footnote{Additional materials can be found at \url{https://srl-ethz.github.io/SERNF/}.}
\end{abstract}


\vspace{-0.5\baselineskip}
\begin{figure}[!h]
  \centering
  \includegraphics[width=0.82\linewidth]{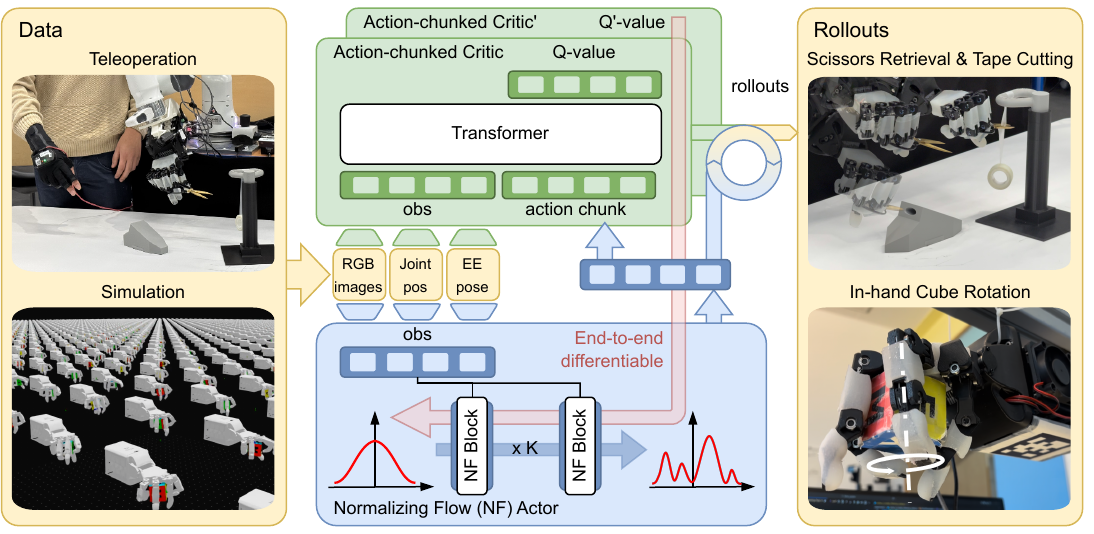}
  \caption{Overview of \algoname (\textbf{S}ample-\textbf{E}fficient \textbf{R}einforcement learning with \textbf{N}ormalizing \textbf{F}lows). \algoname pairs a conditional normalizing-flow actor (expressive, exact likelihood, single-pass inference) with an action-chunked critic. Initialized from real-world teleoperation or simulation-trained policies, \algoname achieves high real-world performance under a limited interaction budget.}
  \label{fig:overview}
\end{figure}
\vspace{-0.5\baselineskip}

\input{sections/introduction}
\input{sections/related_work}
\input{sections/background}
\input{sections/method}
\input{sections/experiments}
\input{sections/results}

\input{sections/conclusion}

\acknowledgments{This work was supported by the Swiss National Science Foundation (SNSF) Project Grant No.~200021\_215489 and the SDSC Grant C22-08. We also acknowledge the ETH AI Center and the NVIDIA Academic Grant Program for providing computational resources.}

\newpage
\bibliography{references}

\clearpage
\appendix
\input{sections/appendix}

\end{document}

%% file: sections/introduction.tex
\section{Introduction}
\label{sec:introduction}
\vspace{-7pt}
Modern visuomotor policies can imitate dexterous manipulation when trained on
large, diverse robot datasets, but this scaling recipe breaks down precisely
where dexterity matters most. Contact-rich tasks such as scissor retrieval,
tape cutting, and in-hand object rotation are hard to teleoperate, require
expert skill, and are sensitive to small errors in contact, timing, and
perception. As a result, real-world policies often ``almost'' work: they
produce plausible behavior yet fail under modest shifts in dynamics,
hardware calibration, camera geometry, or task setup. Reinforcement learning (RL)
fine-tuning is the natural way to turn these near-solutions into reliable
controllers using limited additional interaction, but standard RL
fine-tuning does not fit the policy class now favored for robotics:
high-capacity, multimodal policies that generate temporally extended action
chunks.

Diffusion and flow-matching policies are used as action-generation
models in modern imitation-learning and vision-language-action systems
because they capture rich, multimodal action distributions and naturally
support chunked control. However, they expose two barriers to direct
off-policy RL fine-tuning. First, their action likelihoods are generally
intractable or only available through surrogate objectives, making
conservative likelihood-based updates difficult; and because action
generation unrolls through multiple denoising steps, \emph{end-to-end}
critic-guided fine-tuning would require differentiating through the full
denoising chain, which is memory-intensive and unstable --
on top of inflating inference latency and thus limiting real-time control
rates. Second, standard per-step critics estimate value at a finer
temporal scale than the policy actually controls: the actor commits to
an action chunk, while the critic evaluates individual steps. This
misalignment weakens credit assignment, increases variance, and makes
critic-guided updates unreliable for long-horizon dexterous manipulation.
Our method targets these two bottlenecks directly: \emph{tractable
likelihoods and single-pass inference} for expressive multimodal
policies, and \emph{value estimation aligned with chunked execution}.

We propose \textbf{\algoname}, a \textbf{S}ample-\textbf{E}fficient
\textbf{R}einforcement learning framework with
\textbf{N}ormalizing-\textbf{F}low policies for chunked, multimodal
visuomotor control under limited real-world interaction. \algoname represents
action chunks with a conditional normalizing-flow policy, preserving the
expressiveness needed for multimodal manipulation while providing exact
action likelihoods needed for RL. It pairs this actor with an action-chunked
critic that evaluates entire action sequences, aligning value estimation
with the temporal unit actually executed by the robot. This combination
enables stable critic-guided improvement from both offline data and online
rollouts.

We evaluate on three real-world dexterous manipulation tasks chosen to span
three dominant deployment regimes:
(i) \emph{scissors retrieval and tape cutting} from a small set of human
teleoperation demonstrations -- the common imitation-learning workflow for
contact-rich tasks;
(ii) \emph{in-hand cube rotation} from a simulator-trained teacher distilled
into our architecture -- the mainstream sim-to-real RL workflow for
dexterous hands;
(iii) \emph{rubber-duck pick-and-place} from very few demonstrations and
online rollouts -- short-horizon, multi-pose grasping.
All three are sensitive to small contact and timing errors and serve as
strong stress tests for sample-efficient fine-tuning.

The main contributions are:
  \textbf{(a)} We introduce an RL fine-tuning method for real-world visuomotor
  control that combines \emph{normalizing-flow} policies with an
  \emph{action-chunked critic}, supporting exact likelihood-based
  conservatism and single-pass inference.
  \textbf{(b)} We describe a practical four-stage training recipe
  (IL $\to$ critic warm-up $\to$ offline RL $\to$ online RL) for limited
  on-robot data.
  \textbf{(c)} We empirically demonstrate that \algoname improves scissors cutting
  from $0.16$ to $0.84$ with only 60 additional rollouts -- and on this
  task outperforms a diffusion-policy RL baseline (DPPO), which fails to
  improve over its IL initialization in our regime. \algoname additionally
  reaches $6.25$\,RPM cube rotation in 105\,min of real-world data and
  lifts duck pick-and-place from $0.75$ to $0.87$ with 30 teleoperated and 15 additional online
  rollouts.

%% file: sections/related_work.tex
\section{Related Work}
\label{sec:related_work}
\vspace{-7pt}
\rkkin{\textbf{Related-work strategy (rkk):} See original draft for the comparison-table suggestion.}

\paragraph{Policy optimization for generative policies.}
Diffusion and flow-matching policies enable expressive, multimodal action
distributions for imitation
learning~\citep{black2024pi_0,Chi2023DiffusionPolicy,ze20243DDiffusion}, but
their use in RL is limited by intractable action likelihoods, which hinder
conservative or likelihood-regularized fine-tuning.
Recent work connects diffusion- or flow-based models with policy
optimization~\citep{anonymous2025policyflow,intelligence2025pi,McAllister2025FPO,Ren2025DPPO,Zhang2025ReinFlow}.
Among these, DPPO~\citep{Ren2025DPPO} is the closest to a real-world,
diffusion-RL system; we compare against it directly in
Sec.~\ref{sec:results_scissors} and find that its on-policy nature limits
sample efficiency in our setting. FPO~\citep{McAllister2025FPO} and
ReinFlow~\citep{Zhang2025ReinFlow} are only validated in simulation. In
contrast, we focus on \emph{normalizing flows}, which retain comparable
expressiveness while providing exact likelihoods and single-pass inference,
enabling stable and principled RL-based fine-tuning.
\vspace{-7pt}
\paragraph{Normalizing flows for RL and IL.}
Normalizing flows are effective policy representations for reinforcement and
imitation learning~\citep{Ghugare2025NFCapable,Lind2025NFCapableVisuomotor,tarasov2025nina},
but their application to real robotic systems remains limited. Prior
real-world deployments are restricted to on-policy
settings~\citep{khader2021learning}, while most existing work focuses on
simulation~\citep{akimov2022let,chao2024maximum,mazoure2020leveraging}. Our
work leverages NF policies for visuomotor control and focuses on stable,
conservative fine-tuning on real hardware. To our knowledge, this is the
first off-policy, visuomotor RL fine-tuning of a likelihood-based multimodal
generative policy on real robotic hardware.
\vspace{-7pt}
\paragraph{Temporal abstraction and action chunking.}
Temporally extended action sequences are now standard in robotic control,
enabling real-time execution and reduced feedback
frequency~\citep{black2024pi_0}. Prior work shows that aligning critics with
action chunking improves learning efficiency and credit
assignment~\citep{Li2025QChunking}.
We extend these ideas by integrating action-chunked critics into a real-world
fine-tuning pipeline which was never done before.
\vspace{-7pt}
\paragraph{Offline + online RL.}
Real-world robot learning commonly combines offline initialization with
limited online fine-tuning.
AWAC~\citep{nair2020awac} and related methods enable off-policy adaptation
but lack explicit conservatism in highly multimodal action spaces.
Subsequent work studies the offline-to-online
transition~\citep{ball2023efficient,nakamoto2023cal}, trading off
conservatism and exploration at the cost of increased algorithmic complexity.
Systems approaches such as IBRL and SERL emphasize the practical integration
of demonstrations, off-policy RL, and tooling for real
robots~\citep{hu2023imitation,Luo2024SERL}, while extreme online-only
efficiency has been demonstrated in locomotion from
scratch~\citep{smith2022walk}.
In contrast, \algoname targets sample-efficient real-world fine-tuning of
pretrained visuomotor policies under multimodal action distributions and
sparse rewards, by combining exact-likelihood generative policies with
chunk-aligned value learning.

%% file: sections/background.tex
\section{Background}
\label{sec:background}
\vspace{-7pt}
\rkkin{\textbf{Background section critique (rkk):} Three issues to fix before submission. (1) The \texttt{resizebox} hack on three equations is a clear sign the equations are too wide for one column -- this looks rough at print resolution. Either re-derive in a more compact notation (drop $\substack{\ldots\sim\mathcal{D}}$ subscripts, define a shorthand like $\mathbb{E}_\mathcal{D}$) or move the long equations to a single \texttt{figure*}/\texttt{equation*} spanning two columns. (2) Heavy notation is fine but the section currently re-derives standard offline-RL machinery without a clear ``what is non-standard here'' callout -- shorten by 30\% and use the saved space for a synergy paragraph between NF + chunked critic. (3) Many lines of commented-out duplicated content -- delete before submission.}

\paragraph{MDP with observations and action chunks.}
We consider an episodic, discounted MDP
$\mathcal{M} = (\mathcal{S}, \mathcal{A}, P, r, \gamma, T)$ with sparse rewards,
in which the agent receives observations $o_t \in \mathcal{O}$ and selects an
\emph{action chunk} $\mathbf{a}_k = a_{t:t+H}$, executed open-loop over the next
$H$ environment steps.
Following~\citet{Li2025QChunking}, we use an action-chunked value
$Q_{H\text{-step}}(s_t, a_{t:t+H})$ that propagates returns over an entire
chunk, better matching modern imitation- and visuomotor-learning frameworks
that generate temporally extended action sequences.
\vspace{-7pt}
\paragraph{Offline RL with multi-step bootstrapping.}
In offline RL~\citep{levine2020offline}, we assume a fixed dataset
$\mathcal{D} = \{(s,a,s',r)\}$ collected by an unknown behavior policy
$\pi_\beta$. We train a critic $Q_\phi$ via temporal-difference learning with
$H$-step bootstrapping and a slowly updated target network $Q_{\bar\phi}$,
\begin{equation}
\mathcal{L}_Q(\phi) = \mathbb{E}_{\mathcal{D}}\Big[\big(Q_\phi(s_t, a_{t:t+H})
- {\textstyle\sum_{\tau=1}^H} \gamma^{\tau} r_{t+\tau}
- \gamma^H V_{\bar\phi}(s_{t+H})\big)^2\Big],
\end{equation}
with $V_{\bar\phi}(s_t)= Q_{\bar\phi}(s_t, \mathbf{a}_k)$ and
$\mathbf{a}_k \sim \pi_\theta(\cdot\mid s_t)$. The actor is trained to
maximize the critic while staying close to $\pi_\beta$:
$\theta = \arg\max_{\theta} \mathbb{E}_{\mathcal{D}}\mathbb{E}_{\mathbf{a}\sim\pi_\theta}[Q_{\bar{\phi}}(s_t,\mathbf{a})]$
subject to a divergence constraint $D(\pi_\theta, \pi_\beta) < \epsilon$. This
formulation requires a tractable likelihood $\log\pi_\theta(\mathbf{a}\mid o)$
for any divergence-based regularizer -- the property that motivates a
normalizing-flow policy.\rkk{$\tau$ is also used as a generic time index in the same equation -- pick a different symbol (e.g. $\rho_{\text{Polyak}}$) to avoid collision.}
\vspace{-7pt}
\paragraph{Conditional normalizing flows.}
Normalizing flows (NFs)~\citep{rezende2015variational} represent expressive,
multimodal distributions over high-dimensional action
chunks~\citep{tarasov2025nina,Lind2025NFCapableVisuomotor} and admit stable RL
training~\citep{Ghugare2025NFCapable}. A latent $z_0 \in \mathbb{R}^d$ sampled
from a base distribution $p_0$ is transformed through a sequence of invertible
mappings $f_k(\cdot;c)$ conditioned on $c\in\mathbb{R}^m$ (typically an
encoded observation), $z_K = f_\theta(z_0;c)$, with conditional log-density
\begin{equation}
\label{likelihood}
\log p_\theta(z_K \mid c) = \log p_0(z_0) - \sum_{k=1}^K \log\Big|\det \tfrac{\partial f_k(\cdot;c)}{\partial z_{k-1}}\Big|.
\end{equation}

Crucially, NF inference requires only a \emph{single forward pass} for both
sampling and likelihood evaluation, in contrast to diffusion/flow-matching
policies which require many denoising steps. 
Critic-guided \emph{end-to-end} fine-tuning of an NF actor only
backpropagates through one forward pass, whereas the diffusion case
would have to differentiate through the full denoising chain ---
memory-intensive and unstable under realistic compute budgets \citep{park2025flowQL}. This combination
of expressiveness, exact likelihood, single-pass differentiable training,
and fast inference is what enables conservative regularization toward
$\pi_\beta$ during offline and online fine-tuning while preserving
real-time control rates.
\rkk{Add one sentence on what NFs trade off vs. diffusion: NF requires invertible architectures (limiting expressivity per layer) and grows with $K$. The reader should leave Background knowing both the win (exact likelihoods) and the cost (architectural restrictions). Honesty about trade-offs strengthens the paper.}

%% file: sections/method.tex
\section{Method}
\vspace{-7pt}
\label{sec:method}
\rkkin{\textbf{Method section critique (rkk):} The four-stage pipeline is solid but the prose does not give intuition for WHY each stage exists. Add a one-paragraph ``training pipeline at a glance'' before the subsections, with a small inline diagram (or use an existing one) showing the four stages and what data each consumes. Right now the reader has to assemble the picture from Algorithm 3 and disjointed paragraphs.}


\subsection{Setting}
\vspace{-7pt}
We learn a visuomotor policy $\pi_\theta(\mathbf{a}\mid o)$ deployed on a real
robot. At decision step $k$, the policy samples an action chunk
$\mathbf{a}_k = (a_{k,0},\dots,a_{k,H-1})$ of horizon $H$, executed open-loop
with low-level stabilization. Observations $o_t$ include camera images, robot
proprioception, and a small number of unexecuted actions from the previous
chunk; this enables real-time chunking
(RTC)~\citep{black2025real} and, in our experiments, also improves
performance by providing temporal context.
\vspace{-7pt}
\subsection{Normalizing-flow policy}
\vspace{-7pt}
We parameterize $\pi_\theta(\mathbf{a}\mid o)$ as a conditional normalizing
flow, enabling efficient sampling, end-to-end differentiation and exact
log-likelihood evaluation -- the property critical for conservative
regularization toward an initial policy during fine-tuning. Following
NinA~\citep{tarasov2025nina}, each transformation $f_k(\cdot; c)$ is a
transformer-based coupling layer similar to
RealNVP~\citep{dinh2016density,kolesnikov2024jet}.
Ground-truth actions are normalized to $[-1,1]$, perturbed with small Gaussian
noise $\mathcal{N}(0,\sigma_{\text{noise}}^2)$ (helpful for stable
NF training~\citep{Ghugare2025NFCapable,tarasov2025nina}) and passed through
an element-wise $\tanh^{-1}$ for bounded support.

Each flow block operates on a latent
$z_k = [\mathbf{z}_1,\dots,\mathbf{z}_H]$ partitioned into two equal-sized
subsets $x_{k_1},x_{k_2}$. Conditioned on $c$, $x_{k_1}$ is processed by
stacked self-attention plus cross-attention with $c$, producing $(s,b)$ that
parameterize an affine transform
$y_2 = \exp(\tanh(s))\odot x_{k_2} + b$ while $y_1=x_{k_1}$. Concatenation
yields $z_{k-1}$, ensuring invertibility and a tractable Jacobian. The
schematic (Fig.~\ref{fig:network}) and full pseudocode (Alg.~\ref{alg:nf})
are in Appendix~\ref{app:algorithms}.
At inference we sample latents from $\mathcal{N}(0,\sigma_{\text{sample}}^2 I)$ with
$\sigma_{\text{sample}} \le 1$ to obtain higher-likelihood actions, a trick
shown to improve performance in NF policies~\citep{Lind2025NFCapableVisuomotor}.
Unlike diffusion- or flow-matching-based policies, a single NF forward pass
suffices for both sampling and likelihood evaluation; this enables real-time
control rates that we found cannot be matched by diffusion-policy baselines
(latency $>$0.5\,s in our setup).
\vspace{-7pt}
\subsection{Action-chunked critic}
\vspace{-7pt}
Following~\citet{Li2025QChunking}, we learn a critic aligned with the chunked
control interface,
\begin{equation}
Q_\phi(o_k, \mathbf{a}_k) \approx
\mathbb{E}\Big[{\textstyle\sum_{i=0}^{H-1}} \gamma^i r_{k,i} + \gamma^H V_\phi(o_{k+1})\Big],
\end{equation}
where $o_{k+1}$ is the next decision-boundary observation. Inspired by recent
value-learning stability results~\citep{farebrother2024stop,tarasov2024value},
we parameterize the critic categorically (HL-Gauss~\citep{imani2018improving})
and train with cross-entropy.
\vspace{-14pt}
\subsection{Algorithm}
\label{sec:method_algo}
\vspace{-7pt}
Our approach -- \textbf{S}ample-\textbf{E}fficient
\textbf{R}einforcement learning with \textbf{N}ormalizing \textbf{F}low
(\textbf{\algoname}) -- consists of four stages: imitation learning,
critic warm-up, full offline RL, and online fine-tuning. Action
selection at inference samples $N_\pi$ candidate chunks from $\pi_\theta$ and
picks the highest-Q chunk. Full pseudocode for the training pipeline,
the NF forward/inverse pass, and the inference-time chunk selection is
in Appendix~\ref{app:algorithms} (Alg.~\ref{alg:softflow},
Alg.~\ref{alg:nf}, Alg.~\ref{alg:action-selection}).
\vspace{-7pt}
\paragraph{I. Policy Initialization.}
We initialize $\pi_\theta$ by imitation learning on a filtered dataset of
successful trajectories $\mathcal{D}_{\text{demo}}$ using the NF
log-likelihood loss
\begin{equation}
\mathcal{L}_{\mathrm{IL}}
=
\mathbb{E}_{(o_t,\mathbf{a}^*_{t:t+H}) \sim \mathcal{D}_{\mathrm{demo}}}
\big[ -\log \pi_\theta(\mathbf{a}^*_{t:t+H}\mid o_t) \big]
\label{eq:loss_il_nf}
\end{equation}
yielding $\pi_{\theta_0}$. Best-of-$N$ sampling under
$\pi_{\theta_0}$ -- straightforward with exact likelihoods, awkward with
diffusion -- already provides a small improvement at
inference~\citep{Lind2025NFCapableVisuomotor}.
\vspace{-7pt}
\paragraph{II. Offline critic warm-up.}
\label{sec:method_offline_warmup}
Because the critic is randomly initialized, we first warm it up under
$\pi_{\theta_0}$ before policy updates, using the target
$\sum_{t=0}^{H-1}\gamma^t r_{k,t} + \gamma^H(1-d) Q_{\bar{\phi}}(o_{k+1},\hat{a}_{k+1})$
with $\hat{a}_{k+1}\sim\pi_{\theta_0}$. The warm-started critic can already be
used to rank candidate chunks (Alg.~\ref{alg:action-selection}).
\vspace{-7pt}
\paragraph{III. Full offline RL.}
\label{sec:method_full_offline}
We then fine-tune $\pi_\theta$ with a full offline actor--critic objective
regularized toward the data distribution, following
TD3+BC~\citep{fujimoto2021minimalist}:
$\max_\theta \mathbb{E}[Q_\phi(o,a)] - \lambda\,\mathcal{L}_{\text{IL}}$.
This simple formulation matches or outperforms more complex
alternatives~\citep{tarasov2023revisiting,tarasov2023corl,Ghugare2025NFCapable}
and adapts naturally to offline-to-online fine-tuning by reducing $\lambda$
during online interaction~\citep{beeson2022improving}.
\vspace{-7pt}
\paragraph{IV. Online RL fine-tuning.}
\label{sec:method_online}
Following~\citet{ball2023efficient}, we mix offline and replay data at a
fixed ratio $\rho$ and apply imitation regularization only to the offline
portion; all other components remain unchanged.
\vspace{-7pt}
\subsection{Implementation Details}
\vspace{-7pt}
Most hyperparameters are tuned in simulation
(Appendix~\ref{app:robomimic_results}) and transferred to real-world tasks
without modification. The policy is a conditional normalizing flow with 16
attention-based coupling blocks of width 256; chunk length is $H=10$ in all
experiments. Action noise has $\sigma_{\text{noise}}\!=\!0.05$ (sim) and
$0.01$ (real), consistent with~\citet{tarasov2025nina}; inference uses
$\sigma_{\text{sample}}\!=\!0.7$. We train with two critics, BC-regularization
coefficient $\lambda=0.1$, and HL-Gauss configuration
from~\citet{tarasov2024value}. The offline/online mixing ratio is $\rho=1$ for
all tasks. Additional details, including ablations on chunk length and
$\lambda$, are in Appendix~\ref{app:hyperparameters}.
\vspace{-7pt}

%% file: sections/experiments.tex
\section{Experiments}
\label{sec:experiments}
\vspace{-7pt}
\rkkin{\textbf{Experiments section critique (rkk):} The simulation results, real-world setups, and baselines are all in this section, but the structure is hard to follow because results are split across Experiments and Results sections. Consider merging them, OR clearly mark Experiments as ``setup only''.}

We evaluate \algoname on three real-world dexterous-manipulation tasks chosen
to cover three deployment regimes: (i) \emph{Scissors retrieval and tape
cutting}, a precise multi-stage task initialized from human teleoperation
demonstrations (full IL $\to$ offline RL $\to$ online RL pipeline);
(ii) \emph{In-hand cube rotation}, a continuous palm-down dexterous task
initialized from a simulator-trained teacher (distillation $\to$ online RL);
and (iii) \emph{Rubber-duck pick-and-place}, a short-horizon multi-pose grasp
task that stresses behavior diversity with a tiny dataset (IL $\to$ offline RL $\to$ online RL).
We additionally
validate hyperparameters and the offline-RL stage on the RoboMimic
benchmark; the simulation table is reported in
Appendix~\ref{app:robomimic_results}.
\vspace{-7pt}

\subsection{Scissors Retrieval \& Tape Cutting}
\label{sec:exp_scissors}
\vspace{-7pt}
This task features highly multimodal action distributions and is hard to
simulate due to complex contact. We test whether \algoname can fine-tune an
imitation-initialized policy under a limited real-world data budget.
The robot grasps a pair of scissors from a case and cuts a tape suspended
from a support. The setup is a 7-DoF Franka Emika Panda equipped with a
dexterous ORCA hand~\citep{christoph2025orcaopensourcereliablecosteffective};
the arm is controlled in end-effector pose space (quaternion + position) for
a 24-DoF total action space. The task is precise -- the finger holes are
barely larger than the robot fingertips -- and the cut is performed in
mid-air. Three RGB cameras provide observations: two egocentric
(wrist-mounted) and one external. Hardware photos and additional setup
details are in Appendix~\ref{app:setup_scissor}.
\vspace{-7pt}
\paragraph{Policy initialization.}
We collect 121 teleoperated demonstrations via motion-capture gloves; only 71
fully succeed at both grasp and cut. The IL policy is trained on the
successful subset, while the failure trajectories are reused during
offline-RL fine-tuning -- a real-world realism that IL alone cannot exploit.
\vspace{-7pt}
\paragraph{Implementation.}
Three RGB views are encoded independently by a frozen DINOv2
backbone~\citep{oquab2023dinov2} and concatenated.
Rewards are sparse and manually annotated: $r=1$ for retrieving the scissors,
$r=1$ for a successful cut. We run 5{,}000 steps of critic warm-up
(Sec.~\ref{sec:method_offline_warmup}), 1{,}000 steps of full offline RL
(Sec.~\ref{sec:method_full_offline}), then 6 online RL iterations of 10
rollouts each (60 online trajectories total), with 500 gradient steps per
iteration.
\vspace{-7pt}
\paragraph{Evaluation protocol.}
We evaluate two-stage success rates (grasp / cut). We evaluate over a test grid of 5 tape positions $\times$ 3
scissors positions (15 configurations), in addition to the we test using an additional
10-configuration set (2 tape $\times$ 5
scissors) to strengthen statistical power. Aggregated success rates are reported in
Sec.~\ref{sec:results_scissors}; the test configurations are illustrated
in Appendix~\ref{app:test_configs_scissor}. An episode terminates on
successful cut, dropped scissors, unsafe configuration, or a 120\,s timeout.
\vspace{-7pt}
\subsection{In-hand Cube Rotation}
\label{sec:exp_cube}
\vspace{-7pt}
This task represents the common sim-to-real pipeline in dexterous in-hand
manipulation: it is highly dynamic and effectively impossible to teleoperate
reliably due to latency, retargeting errors, and the absence of haptic
feedback. We test whether \algoname can bridge the sim-to-real gap with
limited real-world fine-tuning.
The ORCA hand is operated palm-down, with no external support. States are
hand joints plus cube pose;
actions are desired joint commands (17-DoF). An RGB camera below the hand
provides cube observations. Pose estimation uses a custom CNN initialized
from Mask R-CNN, followed by a PnP solver as
in~\citep{handa2023dextreme}; details in Appendix~\ref{app:cube_pose}.
\vspace{-7pt}
\paragraph{Policy initialization.}
We pretrain a teacher with PPO in IsaacLab with domain
randomization~\citep{nvidia2025isaaclabgpuacceleratedsimulation}, then distill
into the NF student
(Appendix~\ref{app:distillation}). 
\vspace{-7pt}
\paragraph{Implementation \& evaluation.}
After distillation we run online RL directly
(Sec.~\ref{sec:method_online}), motivated by the large sim-to-real gap of
the simulated trajectories. Reward is sparse: $r=1$ per $90^\circ$ rotation.
We perform 7 online iterations of 15\,minutes each (1{,}000 gradient steps
per iteration), evaluating with continuous 15\,min rollouts and measuring
average rotations per minute (RPM) and cumulative rotation per trajectory.
\vspace{-7pt}
\subsection{Rubber-Duck Pick-and-Place}
\label{sec:exp_duck}
\vspace{-7pt}
To probe \algoname under a third regime -- short-horizon, multi-pose
grasp -- we introduce a real-world pick-and-place task: a rubber duck must
be grasped from varied initial poses (lying, upside-down, on edges, with
random in-plane rotations) and placed into a nearby bowl. We collect 30
teleoperated demonstrations (all successful), run the full
IL\,$\to$\,offline-RL\,$\to$\,online-RL pipeline, and use 15 additional
online rollouts for the online stage. Evaluation uses 2 fixed positions
$\times$ 8 orientations (16 trials per checkpoint). Setup details and
rollout snapshots are in Appendix~\ref{app:duck}.
\vspace{-7pt}
\subsection{Baselines}
\label{sec:baselines}
\vspace{-7pt}
We compare against three families of imitation baselines and one RL baseline.
For imitation we evaluate (i) ACT~\citep{zhao2023learning},
(ii) Diffusion Policy~\citep{Chi2023DiffusionPolicy} (denoted \textbf{Diffusion IL}),
and (iii) our normalizing-flow head trained with IL only (\textbf{NF IL}).
For RL, we additionally compare against \textbf{DPPO}~\citep{Ren2025DPPO},
which first pretrains a diffusion policy with imitation learning and then
fine-tunes on-policy in the same robot deployment pipeline as ours. We
reproduce DPPO with the original hyperparameters and our chunk size, RTC, and
rollout protocol. We also attempted an ACT+RL with MSE behavior-cloning
regularization, but it diverged after the critic-warmup stage, likely due to
high sensitivity of the regularization coefficient to architecture and data;
we report this negative result in Appendix~\ref{app:act_rl}. Other recent
flow-matching/NF RL baselines (FPO~\citep{McAllister2025FPO},
ReinFlow~\citep{Zhang2025ReinFlow}) have only been validated in simulation,
and adapting them to real hardware was not feasible within our experimental
budget; we discuss them qualitatively in
Sec.~\ref{sec:related_work}. To probe data-leverage, we also include two
ablations: an IL policy trained on all successful trajectories collected
throughout training (including online rollouts), and an IL policy trained on
the original demonstrations augmented with additional teleoperated data
matched to \algoname's total trajectory count.

%% file: sections/results.tex
\vspace{-7pt}
\section{Results}
\label{sec:results}
\vspace{-7pt}
\rkkin{\textbf{Results section critique (rkk):} Three structural fixes. (1) Open with a 1-sentence ``what we will show'' summary. (2) Each subsection should answer one falsifiable claim. (3) Negative findings (the late-stage grasp dip) -- highlight them as an analysis callout.}

We report results for the three real-world tasks. The headlines are: on the
scissors task, \algoname raises the cutting success rate from $0.16$ (IL) to
$0.84$ (full \algoname) under an expanded 25-configuration evaluation, while
out-performing DPPO -- a strong on-policy diffusion-RL baseline -- which
fails to improve over its IL initialization; on the cube task, \algoname
reaches $6.25$\,RPM after 105\,min of real-world data; and on the duck task,
\algoname raises the success rate from $0.75$ to $0.87$ with only 15
additional online rollouts.
\vspace{-7pt}
\subsection{Scissors Retrieval \& Tape Cutting}
\label{sec:results_scissors}
\vspace{-7pt}
\begin{figure}[t]
\centering
\begin{minipage}[b]{0.33\linewidth}
  \centering
  \includegraphics[width=\linewidth,height=2.6cm,keepaspectratio]{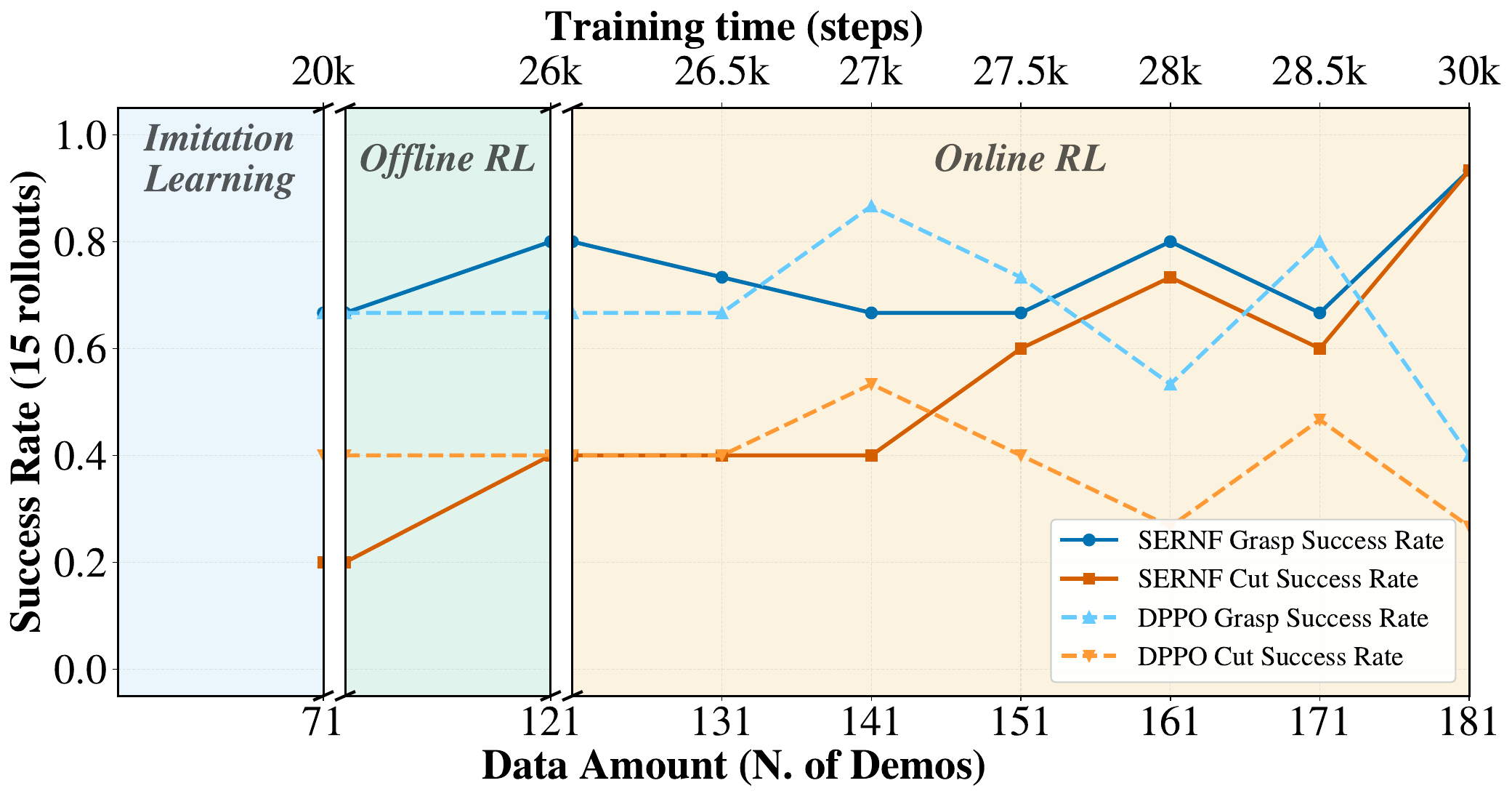}\\
  {\footnotesize (a) Scissors: \algoname vs.~DPPO}
\end{minipage}\hfill
\begin{minipage}[b]{0.33\linewidth}
  \centering
  \includegraphics[width=\linewidth,height=2.6cm,keepaspectratio]{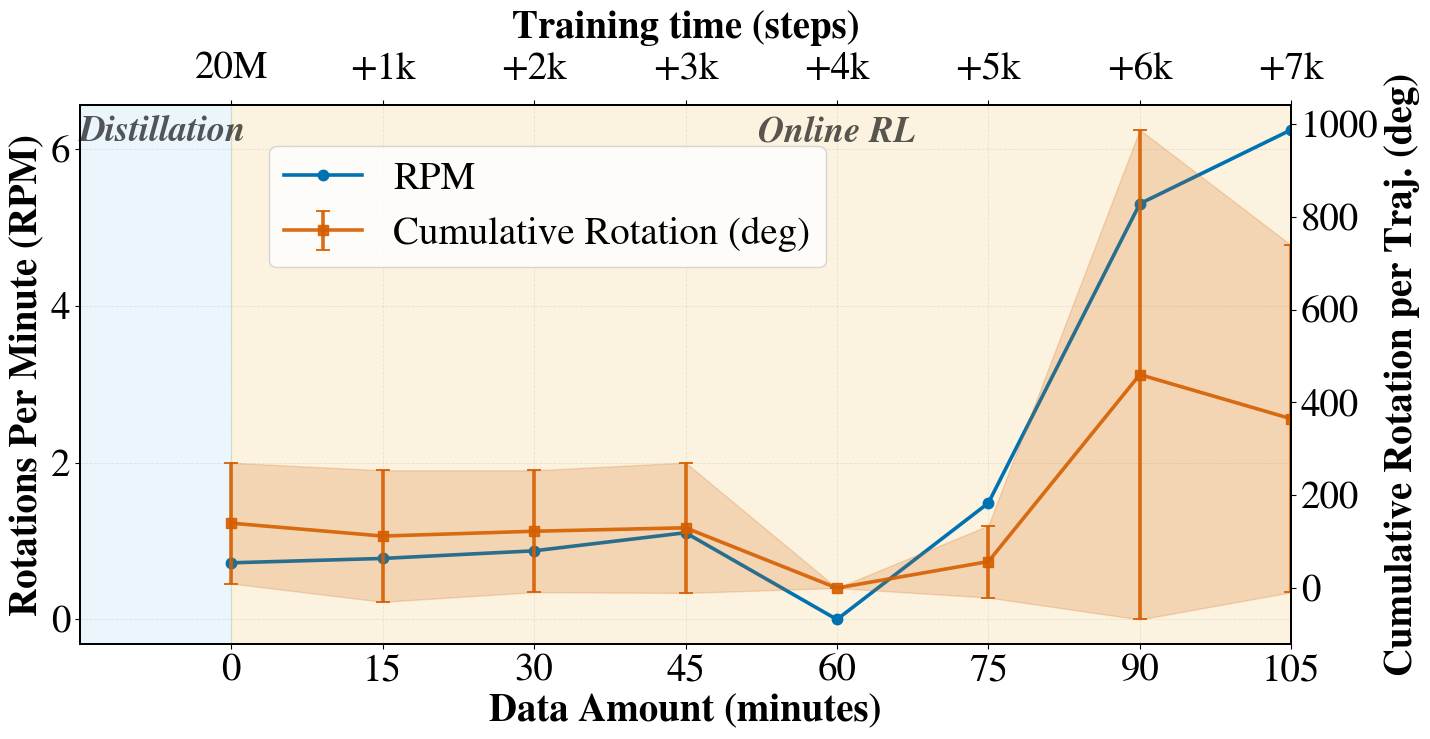}\\
  {\footnotesize (b) Cube rotation}
\end{minipage}\hfill
\begin{minipage}[b]{0.33\linewidth}
  \centering
  \includegraphics[width=\linewidth,height=2.6cm,keepaspectratio]{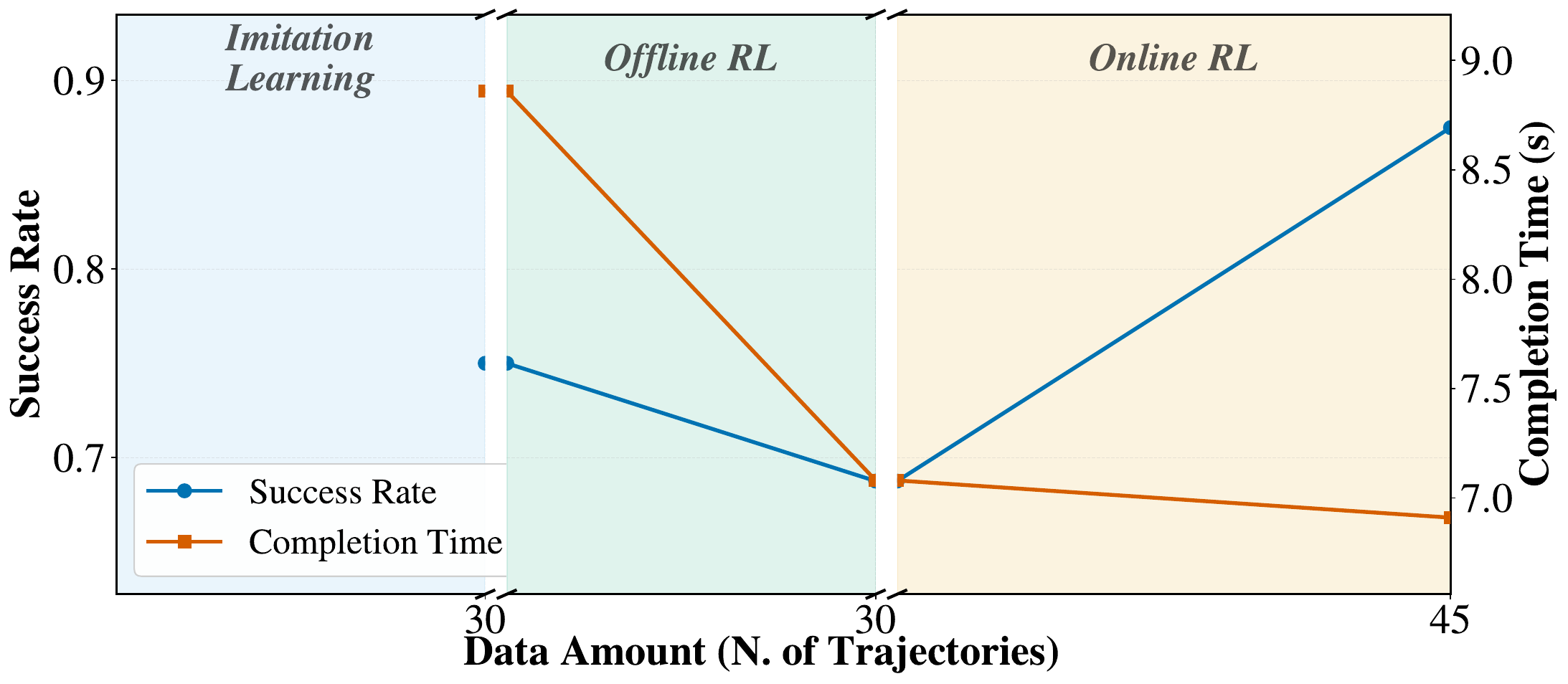}\\
  {\footnotesize (c) Duck pick-and-place}
\end{minipage}
\caption{Per-task training dynamics of \algoname.
\textbf{(a) Scissors}: success rate over the expanded test grid (15
configurations) for \algoname and DPPO. After offline RL, \algoname already
outperforms DPPO overall; online RL closes the cutting gap. DPPO does not
improve over its IL initialization. Aggregated 25 rollouts \algoname: IL
$(0.6\,/\,0.16)$, offline $(0.8\,/\,0.24)$, full $(0.84\,/\,0.84)$.
\textbf{(b) Cube}: rotations per minute (blue) and cumulative rotation per
trajectory (orange) versus real-world data. After 3{,}000 critic warm-up
updates, actor fine-tuning briefly dips, then improves rapidly, peaking at
$6.25$\,RPM with $1.01$ full turns per trajectory after 105\,min of
real-world data.
\textbf{(c) Duck}: success rate (blue) and average completion time on the
intersection of successful rollouts (orange). Offline RL slightly lowers
SR ($0.75\to0.68$) but speeds up successful executions ($8.86\to7.08$\,s);
online RL combines both effects ($0.87$ SR, $6.91$\,s).}
\label{fig:main_results}
\vspace{-10pt}
\end{figure}

\begin{wraptable}{r}{0.52\columnwidth}
\vspace{-6pt}
\caption{Scissors and Duck success rates for \algoname and baselines on
the expanded evaluation grid (15 configurations for scissors, 16 for
duck). Duck completion time is averaged over the intersection of
successful rollouts across the three \algoname stages.}
\label{tab:baselines}
\footnotesize
\setlength{\tabcolsep}{3pt}
\renewcommand{\arraystretch}{0.95}
\begin{tabular}{l|cc}
\toprule
\textbf{Scissors} & \textbf{Grasping} & \textbf{Cutting} \\
\midrule
ACT IL                   & 0.53 & 0.20 \\
Diffusion IL             & 0.66 & 0.40 \\
NF IL                    & 0.66 & 0.20 \\
ACT + offline RL         & 0.00 & 0.00 \\
DPPO~\citep{Ren2025DPPO} & 0.40 & 0.26 \\
\algoname (Offline only) & 0.80 & 0.40 \\
\algoname (Full)         & \textbf{0.93} & \textbf{0.93} \\
\midrule
\textbf{Duck}            & \textbf{Success rate} & \textbf{Compl.~time (s)} \\
\midrule
NF IL                    & 0.75 & 8.86 \\
\algoname (Offline only) & 0.68 & 7.08 \\
\algoname (Full)         & \textbf{0.87} & \textbf{6.91} \\
\bottomrule
\end{tabular}
\vspace{-6pt}
\end{wraptable}

Table~\ref{tab:baselines} and Fig.~\ref{fig:main_results}(a) summarize scissors-task
performance. The three IL policy classes (ACT, Diffusion, NF) achieve
comparable grasping (0.53--0.66) but very low cutting (0.20--0.40). ACT
augmented with our offline-RL stage collapses to 0/0 -- consistent with the
known sensitivity of MSE-regularized RL to multimodal data
(Appendix~\ref{app:act_rl}). DPPO matches its IL initialization (0.40 /
0.26) but does not improve under online fine-tuning, attributable to its
on-policy nature and sample inefficiency in our limited-budget regime. In
contrast, \algoname improves monotonically across stages: offline RL
already lifts grasping from 0.66 to 0.80, and online RL takes both metrics
to 0.93; aggregated across the original and additional test grids (25 trials), IL
$(0.6/0.16)$ $\to$ offline $(0.8/0.24)$ $\to$ full $(0.84/0.84)$.
Augmenting IL with more teleoperation or with on-policy trajectories
collected during online rollouts yields no improvement or hurts performance
(Appendix~\ref{app:il_ablations}), confirming that the gains come from
critic-guided improvement, not extra data. A small grasp-rate dip late in
online training is analyzed in Appendix~\ref{app:failures}.
\vspace{-10pt}

\subsection{In-hand Cube Rotation}

\label{sec:results_cube}
\vspace{-7pt}
\algoname bridges sim-to-real for in-hand cube rotation
(Fig.~\ref{fig:main_results}(b)). The distilled init is fragile -- most
trajectories fail before a full rotation. After critic warm-up and a
brief exploration dip, performance improves rapidly to $1.49$\,RPM at
5k updates and peaks at $6.25$\,RPM at 7k updates with $1.01$ cumulative
turns per trajectory, transitioning from short fragile holds to stable
grasp maintenance with rapid continuous rotation.
\begin{figure}[t]
\centering
\includegraphics[width=\linewidth]{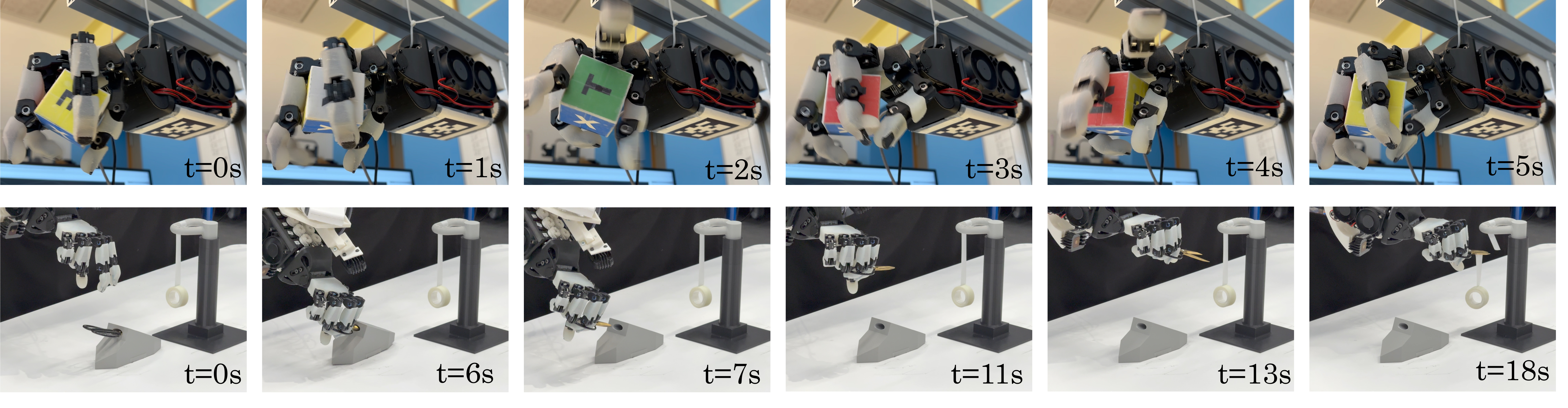}
\vspace{-22pt}
\caption{Qualitative real-world rollouts of \algoname.
\textbf{Top:} successful scissors retrieval and tape cutting.
\textbf{Bottom:} in-hand cube rotation.}
\label{fig:exp_photos}
\vspace{-10pt}
\end{figure}

\vspace{-10pt}
\subsection{Rubber-Duck Pick-and-Place}
\label{sec:results_duck}
\vspace{-7pt}
From just 30 teleoperated demonstrations, NF IL already achieves a strong
$0.75$ SR (Table~\ref{tab:baselines}, bottom). Offline RL on the same
30-trajectory dataset slightly lowers SR to $0.68$ -- expected when
training only on successful trajectories, with no failure signal -- while
changing behavior toward faster but less precise grasps. With just 15
additional online rollouts, online RL then raises SR to $0.87$.
Beyond success, RL accelerates execution: on the intersection of
successful rollouts across stages, average completion time drops from
$8.86$\,s (IL) to $7.08$\,s (offline RL) to $6.91$\,s (online RL)
(Fig.~\ref{fig:main_results}(c)) -- so even at lower SR, offline RL
completes the task substantially faster than IL, and online RL combines
both. Rollout snapshots: Appendix~\ref{app:duck}.

%% file: sections/conclusion.tex
\vspace{-7pt}
\section{Limitations and Future Work}
\vspace{-7pt}
\label{sec:limitations}

\paragraph{Real-world data and safety.}
Although \algoname is designed to be sample-efficient, it still requires
direct interaction with the robot, which entails wear, supervision, and
safety considerations during exploration. Scaling \algoname to many
tasks, to continual on-robot adaptation, or to deployments with weaker
safety guarantees remains an open challenge.
\vspace{-7pt}
\paragraph{Reward design.}
Our experiments rely on sparse, manually annotated rewards. Manual
labeling does not scale and can be poorly aligned for tasks beyond those
we study. Automating reward labeling -- for example with vision-language
models that judge task completion from rollout video -- is a natural
next step toward applying \algoname to broader task sets.
\vspace{-7pt}
\paragraph{Evaluation scope.}
Each real-world checkpoint is evaluated on $15$--$25$ trials per task.
While this is comparable to prior real-world manipulation papers, the
resulting statistical power is limited; per-configuration breakdowns
and additional failure-mode analysis are provided in
Appendix~\ref{app:failures}. Larger evaluation grids and multi-seed
training would further sharpen the conclusions but were not feasible
within our compute and robot-time budget.
\vspace{-7pt}
\paragraph{Initialization and policy class.}
\algoname assumes a competent IL- or distillation-based initialization;
behavior under poor initialization is unstudied, and the NF actor's
invertibility constraints grow in cost with depth and chunk
dimensionality. Promising extensions include integrating \algoname with
large-scale vision--language--action
policies~\citep{intelligence2025pi}, and benchmarking on offline and
offline-to-online RL
suites~\citep{fu2020d4rl,park2024ogbench}.
\vspace{-7pt}
\paragraph{Scaling to bimanual control (preliminary).}
Appendix~\ref{app:biman} reports a preliminary bimanual experiment
($48$-DoF joint action space, two Franka arms with ORCA hands, tube
and tennis-ball pickups). With the same NF capacity as the
single-arm tasks, online RL drives tube pickup from $1/16$ to $12/16$
while ball pickup remains low ($\leq 3/16$); the positive slope on
both sub-tasks and very fast on-policy data collection suggest the
gap is primarily a budget and model-capacity issue.

\vspace{-7pt}
\section{Conclusion}
\label{sec:conclusion}
\vspace{-7pt}
We presented \algoname, a sample-efficient real-world visuomotor
fine-tuning framework that combines a conditional normalizing-flow
actor with an action-chunked critic. The actor preserves the
expressiveness of modern multimodal action heads while exposing exact
log-likelihoods for conservative regularization and enabling real-time
control via single-pass inference. The critic operates over entire
action chunks, aligning value estimation with the temporal unit the
robot actually executes. Across three real-world dexterous tasks,
\algoname raises scissors cutting from $0.16$ to $0.84$ with only 60
additional online rollouts -- outperforming the DPPO diffusion-RL
baseline -- reaches $6.25$\,RPM cube rotation in 105\,min from a
simulator-distilled policy, and lifts duck pick-and-place from $0.75$
to $0.87$ with 15 additional rollouts while reducing execution time
from $8.86$\,s to $6.91$\,s. Together these results suggest that
pairing exact-likelihood generative policies with chunk-aligned value
learning is a practical recipe for RL fine-tuning of multimodal,
contact-rich dexterous manipulation on real hardware.

%% file: sections/appendix.tex

\section{Algorithms}
\label{app:algorithms}

We provide pseudocode for the full \algoname training pipeline
(Alg.~\ref{alg:softflow}), the normalizing-flow forward/inverse pass
(Alg.~\ref{alg:nf}), and inference-time chunk selection
(Alg.~\ref{alg:action-selection}).

\begin{algorithm}[h]
\caption{\algoname training pipeline}
\label{alg:softflow}
\KwIn{Demo data $\mathcal{D}_{\text{demo}}$, offline data $\mathcal{D}$}
\KwOut{Fine-tuned policy $\pi_\theta$ and critic $Q_\phi$}

\Comment{\textbf{Stage I: Imitation learning}}
\ForEach{minibatch $(o,a)\sim \mathcal{D}_{\text{demo}}$}{
  $\mathcal{L}_{\text{IL}} \gets -\mathbb{E}[\log \pi_\theta(a\mid o)]$\;
  $\theta \gets \theta - \eta_\pi \nabla_\theta \mathcal{L}_{\text{IL}}$\;
}

\Comment{\textbf{Stage II: Offline critic warm-up}}
\Repeat{convergence on $\mathcal{D}$}{
  $\hat{a}_{k+1} \sim \pi_{\theta_0}(\cdot\mid o_{k+1})$\;
  $y \gets \sum_{t=0}^{H-1}\gamma^t r_{k,t} + \gamma^H(1-d)Q_{\bar{\phi}}(o_{k+1},\hat{a}_{k+1})$\;
  $\mathcal{L}_Q \gets \text{CrossEntropy}(Q_\phi(o_k,a_k),y)$\;
  $\phi \gets \phi - \eta_Q \nabla_\phi \mathcal{L}_Q$\;
}

\Comment{\textbf{Stage III: Full offline RL}}
\Repeat{convergence on $\mathcal{D}$}{
  Update critic as above\;
  $a_{\pi_\theta} \gets$ Alg.~\ref{alg:action-selection}\;
  Compute $\log\pi_\theta(a_{d}\mid o)$ using Alg.~\ref{alg:nf}\;
  $\theta \gets \theta - \eta_\pi (\mathbb{E}[Q_\phi(o,a_{\pi_\theta})] - \lambda\,\mathbb{E}[\log\pi_\theta(a_{d}\mid o)])$\;
}

\Comment{\textbf{Stage IV: Online fine-tuning}}
\While{interaction budget remains}{
  Collect rollouts using Alg.~\ref{alg:action-selection}\;
  Add data to replay buffer\;
  Update actor and critic with mixed (ratio $\rho$) offline/online data\;
}
\end{algorithm}

\begin{algorithm}[h]
\caption{Normalizing-flow policy over action chunks}
\label{alg:nf}
\KwIn{Observation $o$, base distribution $p_0(z)$, invertible flow $f_\theta(\cdot;\,o)$}

$c \gets \mathrm{Enc}(o)$\;
\textbf{Forward pass (likelihood evaluation):}\;
$z_0 \gets f_\theta(a;\,c)$ \Comment{actions $\rightarrow$ latent}\;
$\log \pi_\theta(a\mid o) \gets \log p_0(z_0) + \log\left|\det J_{f_\theta}(a;\,c)\right|$\;
\textbf{Inverse pass (action generation):}\;
$z_0 \sim p_0(z)$\;
$a \gets f_\theta^{-1}(z_0;\,c)$ \Comment{latent $\rightarrow$ actions}\;
$\log \pi_\theta(a\mid o) \gets \log p_0(z_0) - \log\left|\det J_{f_\theta^{-1}}(z_0;\,c)\right|$\;
\end{algorithm}

\begin{algorithm}[h]
\caption{Action-chunk selection with critic evaluation}
\label{alg:action-selection}
\KwIn{Observation $o_k$, policy $\pi_\theta$, critic $Q_\phi$, number of samples $N_\pi$}
\KwOut{Selected action chunk $a_k$}

\Comment{Sample and score candidate chunks}
\For{$i=1$ to $N_\pi$}{
  $a^{(i)} \sim \pi_\theta(\cdot\mid o_k)$\;
  $q^{(i)} \gets \min_j Q_\phi^{(j)}(o_k, a^{(i)})$\;
}
$a_k \gets \arg\max_i q^{(i)}$\;
\Return{$a_k$}\;
\end{algorithm}

\section{Hardware Setup and Task Details}
\label{app:tasks}

\begin{figure}[h]
     \centering
     \includegraphics[width=0.95\linewidth]{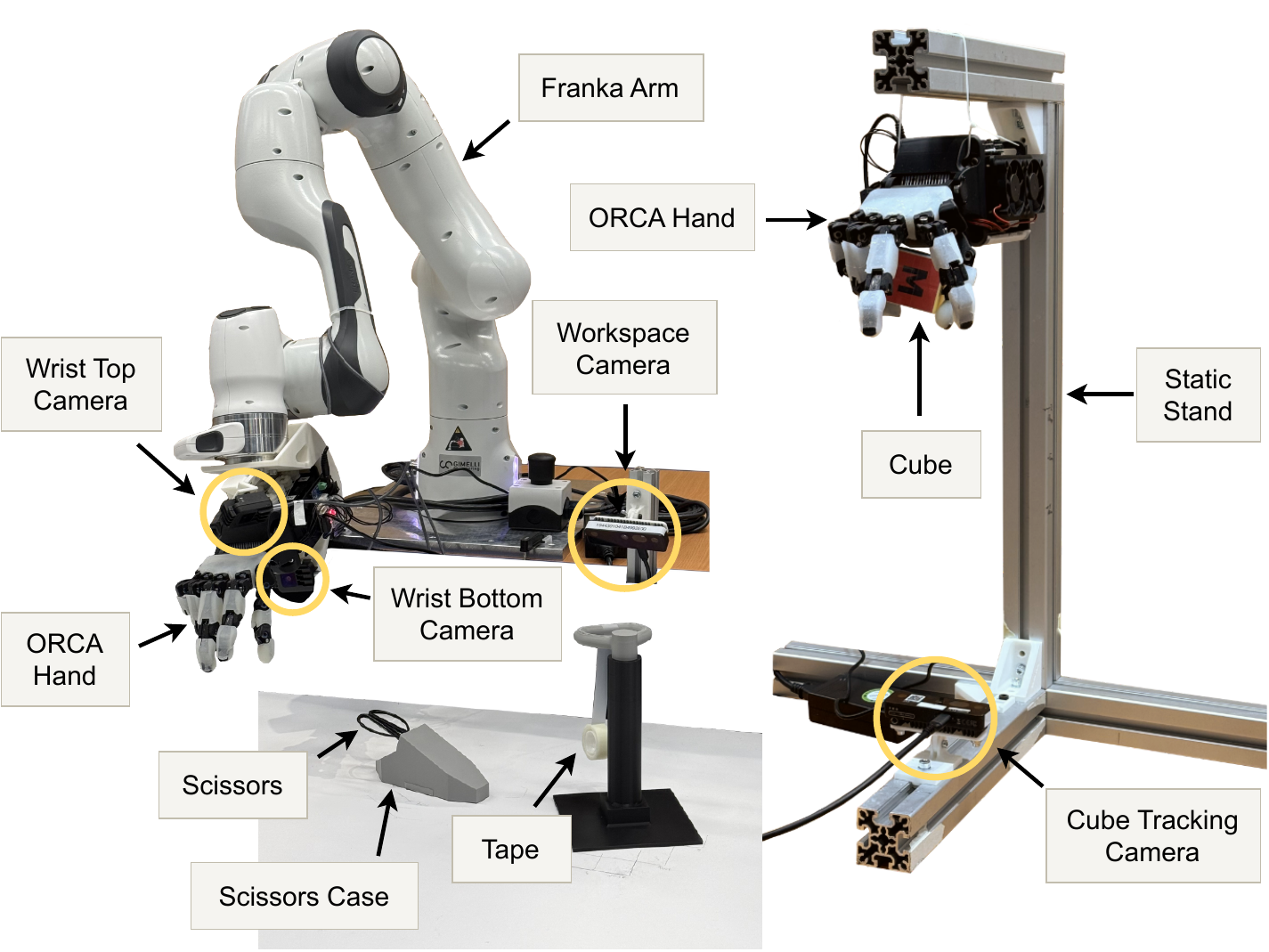}
     \caption{Real-world experimental setup. \textbf{Left:} scissors retrieval
     and tape cutting with a 7-DoF Franka Panda arm and ORCA hand, using two
     wrist-mounted RGB cameras and one external workspace camera.
     \textbf{Right:} in-hand cube reorientation with the ORCA hand under
     continuous palm-down rotations, using single-camera vision-based pose
     estimation.}
     \label{fig:system}
\end{figure}

\subsection{Scissors Retrieval and Tape Cutting}
\label{app:scissor_task}

\paragraph{Experimental setup.}
\label{app:setup_scissor}
We mount the Orca Hand \citep{christoph2025orcaopensourcereliablecosteffective} on a Franka Emika Panda robot using a custom 3D-printed mount with a 60-degree tilt. Two OAK-1 Lite cameras are attached to the mount: one positioned underneath the hand to enable accurate finger placement, and the other facing left to provide visual guidance during scissor manipulation. The third camera is an OAK-D Lite that provides a front view. We use a single pair of craft scissors with a length of 144\,mm and a width of 65\,mm, to cut 3M Scotch Magic Tape, which is suspended freely and allowed to hang naturally. The same pair of scissors is used throughout all experiments.

The Franka low-level controller runs on an Intel NUC at 1K Hz, receiving end-effector pose commands and executing pose impedance control. A separate PC equipped with Intel i9-11900K CPU and Nvidia RTX3090 runs the policy inference at 10Hz and a hand controller that commands the finger motors at 40 Hz. Communication between cameras, controllers, and the policy inference node is handled using ROS 2. The PC, NUC, and Franka controller are connected via Ethernet through a network switch. A ROS 1--ROS 2 bridge is used to interface the Franka controller with the policy inference node. All camera images are cropped and resized to $224\times 224$ before being passed to the policy. The inference policy runs at 10\,Hz.

\paragraph{Teleoperation and data collection.}
\label{app:teleoperation}
Expert demonstrations are collected using Rokoko Smart Gloves in combination with the Rokoko Coil Pro, which together provide finger motion capture and wrist pose estimation in 3D space. Human finger postures are retargeted to the robotic hand joint angles using an energy-based retargeting method \citep{sivakumar2022retargeter}.

All demonstrations attempt to complete the full cutting task from a randomized initial configuration. Operators teleoperate the system with direct visual observation of the scene and make their best efforts to execute the task smoothly. Failed trajectories are not intentionally collected; however, the overall demonstration success rate is approximately 59\%, primarily due to scissor drops, activation of the Franka protective mode, or unintended tilting of the scissor stand. The demonstrations are collected by two operators, each exhibiting distinct teleoperation styles. The duration of the demonstrations ranges from 10 to 60 seconds. The recorded rosbag is sampled to a 10\,Hz sequence with linear interpolation.

Rewards are labeled manually after demonstrations are collected. A sparse reward of 1 is assigned in the following cases:
\begin{itemize}
\item Successful scissor retrieval: the hand securely grasps the scissors and the scissors do not have contact with any objects other than the hand.
\item Successful tape cutting: the tape is fully cut into two separate pieces.
\end{itemize}

\begin{figure}[h]
    \centering
    \includegraphics[width=0.99\linewidth]{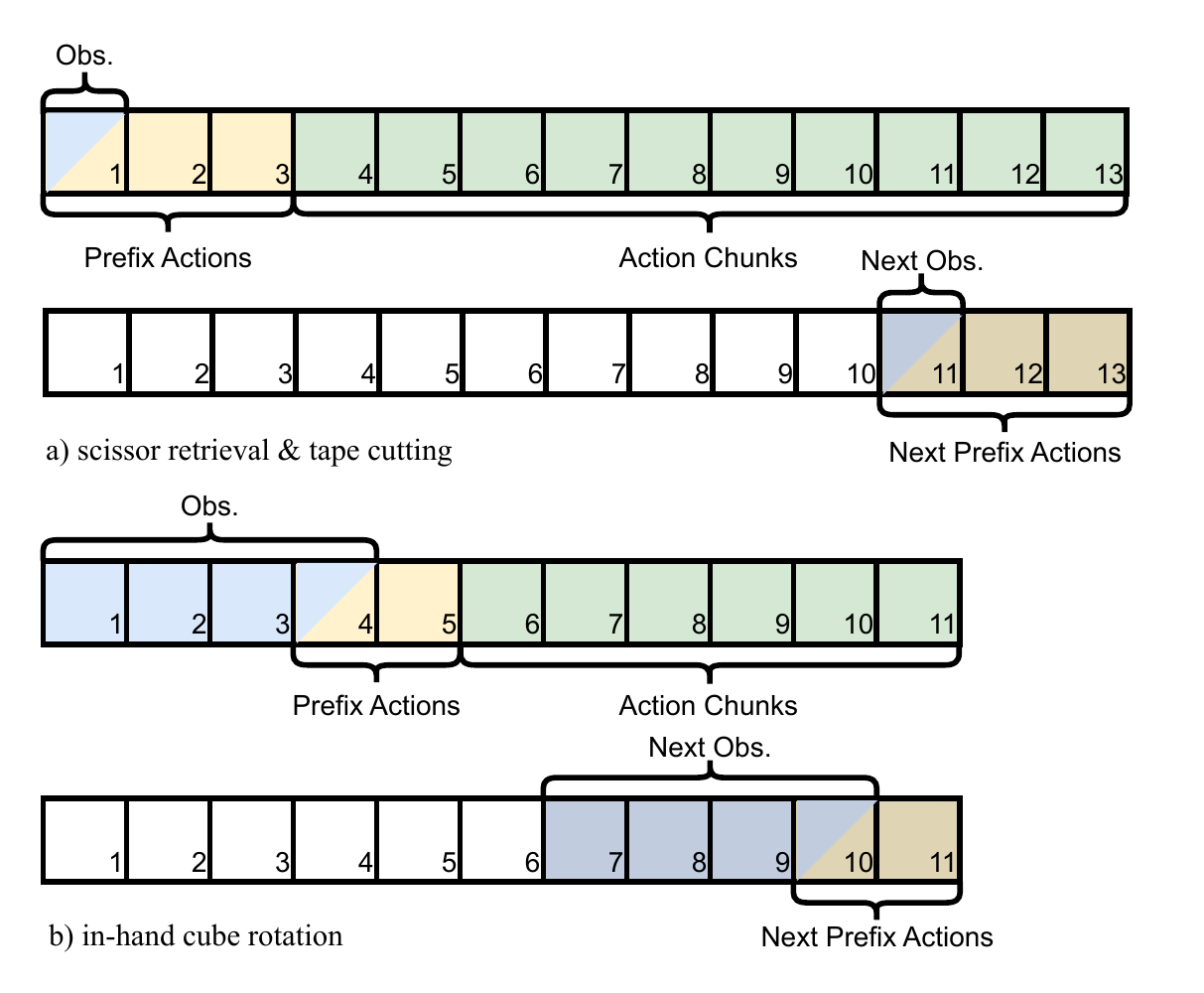}
    \caption{Observation and real-time action-chunking structure for both tasks.
At each decision step, the policy takes as input the current observation together with a sequence of prefix actions, which are previously predicted actions that are queued and ready to be executed, and predicts a temporally extended action chunk.
The figure also illustrates how the next observation and next prefix actions, used for reinforcement learning updates, are defined based on the executed portion of the action chunk.
(a) Scissor retrieval and tape cutting: the policy uses 1 step of observation and 3 steps of prefix actions, and predicts a subsequent chunk of 10 actions.
(b) In-hand cube rotation: the policy uses 4 steps of position observations and 2 steps of prefix actions, and predicts a chunk of 6 future actions.}
    \label{fig:RTC_setups}
\end{figure}

\paragraph{Action chunking and real-time inference.}
\label{app:realtime_inference_scissor}
As shown in Fig.~\ref{fig:RTC_setups}, the policy uses 1 step of observation and 3 steps of prefix actions following the observation and predicts a subsequent chunk of 10 steps of actions. The observations contain the RGB images, the joint positions of the hand, and the end effector pose of Franka. The actions are represented as relative positions of the end effector within the current pose frame.

\begin{figure}[h]
    \centering
    \includegraphics[width=\linewidth]{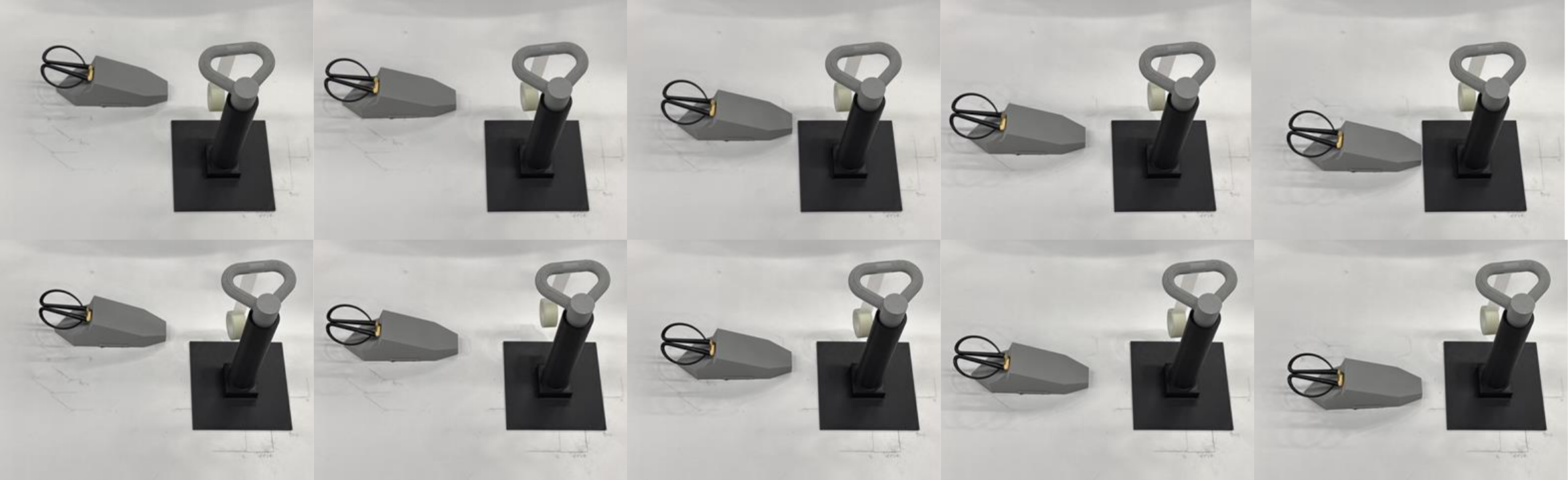}
    \caption{Test configurations for the scissor retrieval and tape cutting task.
The figure shows the 10 predefined test positions used for evaluation, covering variations in the relative placement of the scissors and the tape holder.
The tape length is randomized across trials while ensuring that the tape remains freely hanging and can be successfully cut in all configurations.}
    \label{fig:scissor_test_configs}
\end{figure}

\paragraph{Test configurations.}
\label{app:test_configs_scissor}
We evaluate all policies under a fixed set of predefined test configurations.
Specifically, we define 2 distinct test positions for the scissors and 5 distinct test positions for the tape holder. A total of 10 combinations can be seen in Fig.~\ref{fig:scissor_test_configs}.

For each evaluation episode, a test configuration is sampled from the predefined set. The length of the tape is randomized across trials while ensuring that it remains freely hanging and always reachable and cuttable by the robot. During on-policy reinforcement learning rollouts, the initial configurations are randomized to increase the diversity of collected experience.
All other experimental conditions, including the scissors, tape type, and hardware setup, are held constant across evaluations.

\subsection{In-hand Cube Rotation}
\label{app:cube_task}

\paragraph{Experimental setup.}
\label{app:setup_cube}
We mount the Orca Hand~\citep{christoph2025orcaopensourcereliablecosteffective} horizontally with an OAK-D Lite camera providing the view from the bottom. Both the training and the inference run on a desktop with an NVIDIA RTX 4090 GPU.

\paragraph{Cube pose estimation.}
\label{app:cube_pose}
We use an OAK-D Lite camera mounted approximately 30~cm below the hand holding the cube. To estimate the cube pose, which is provided as input to the policy, we follow the same approach as in~\cite{handa2023dextreme}. We also use the same cube, as its colors and surface features are beneficial for tracking.

A large synthetic dataset is generated using parallel IsaacLab environments. During policy rollouts, images are rendered from a bottom-up viewpoint across 200 parallel environments, and cube corner keypoints are automatically labeled by projecting the 3D cube corners into the image. Using an RTX 5090 GPU, we collected approximately 4\,M images in 2 hours. During data generation, the camera pose, lighting conditions, and object textures are randomized.

\begin{figure}[h]
    \centering
    \includegraphics[width=0.99\linewidth]{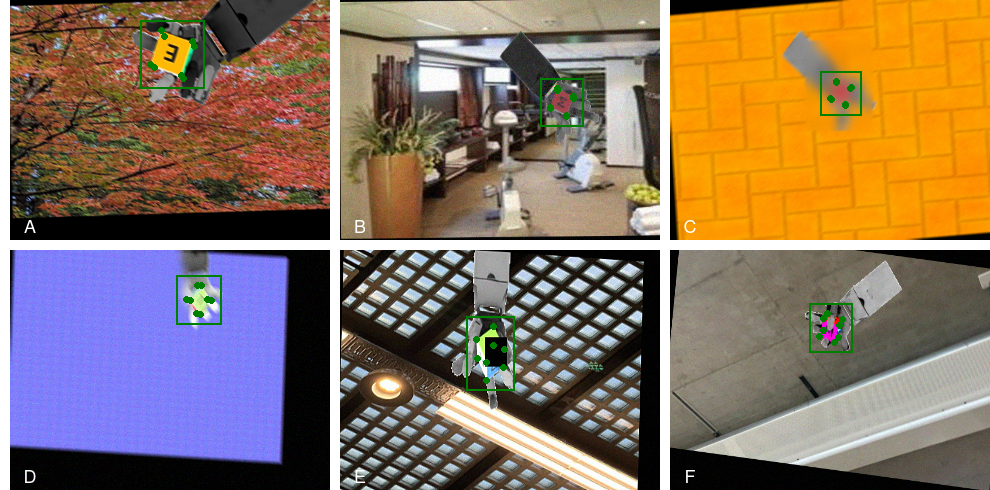}
    \caption{Examples of synthetic training samples. Images are rendered during parallelized IsaacLab training with randomized camera poses, lighting conditions, and hand textures. Additional data augmentations are applied offline, including random cropping, color jitter, random rotation, motion blur, and synthetic occlusions. Backgrounds are randomly sampled from the Places365 dataset~\cite{zhou2017places} (A, B), IsaacSim textures and materials (C, D), and custom in-house images of ceilings and walls (E, F).}
    \label{fig:augmentations}
\end{figure}

We train a Mask R-CNN model, also adopted from~\cite{handa2023dextreme}, consisting of a bounding-box ROI head, a mask head, and a keypoint head on top of the ROI features. The total number of parameters is around 60M. Input images are resized to $240\times 320$. To improve robustness to real-world deployment and severe occlusions, we apply heavy data augmentation, including random cropping (scale 0.7--1.0), small random rotations ($\pm 10^\circ$) with anisotropic stretching, Gaussian blur and noise, random occlusions, and background replacement. Examples of augmented samples are shown in Fig.~\ref{fig:augmentations}.

The network is trained for 200{,}000 steps with a batch size of 16 and a learning rate of $1\times 10^{-8}$. At inference time, the network predicts the cube keypoints, which are then used in a PnP-RANSAC pipeline to estimate the 6-DoF cube pose. The resulting pose is further smoothed using a low-pass Kalman filter for improved stability. The pipeline runs in real time at 16~Hz.

Despite these measures, performance remains challenging due to severe finger occlusions during in-hand rotation and suboptimal lighting conditions. To mitigate pose estimation errors, we apply extensive domain randomization to the cube pose during RL policy initialization, improving robustness to inaccuracies in real-world pose estimates.

\paragraph{Action chunking and real-time inference.}
\label{app:realtime_inference_cube}
As shown in Fig.~\ref{fig:RTC_setups}, at each decision step, the policy receives as input a history of four timesteps of joint positions, cube position, and cube orientation (represented as a quaternion), as well as one timestep of the previous action and the previous joint position command. The goal command is also included as part of the observation.
Conditioned on these observations and a sequence of two prefix actions, the policy predicts an action chunk consisting of six timesteps of target joint positions. The predicted joint positions are smoothed using an exponential moving average with a smoothing coefficient of 0.5.

\paragraph{Teacher policy training.}
\label{app:teacher_policy}
We train the teacher policy fully in simulation using IsaacLab, in an environment that models single-axis spinning of a 45\,mm cube with the Orca hand.

\subparagraph{Simulation setup.}
Each environment instance contains an Orca hand and a rigid cube object placed above a small kinematic platform. The platform provides support for the first 6 seconds in each episode. After the support phase, the platform moves out of the way, allowing the cube to fall freely.
We run 8192 parallel environments.
The simulation runs at 120\,Hz with control frequency at 20\,Hz and a finite-horizon episode length of 32\,s.

\subparagraph{Observations, commands, and actions.}
The teacher uses low-dimensional state observations (no images): normalized hand joint positions and relative joint velocities, cube position and orientation (quaternion), cube linear and angular velocities, the commanded target angular velocity, the previous action, the previous joint-position command, and a counting down of the support of the platform. No noise is added to the observations for teacher policy training.
The action is a target joint-position command (scaled to joint limits) filtered with an exponential moving average (EMA) with coefficient $\alpha=0.5$.
During each episode, we sample a target cube rotation command around the vertical axis with angular velocity up to 1.5\,rad/s.
The command is held for 8--12\,s before resampling.

\subparagraph{Reward shaping.}
The teacher is trained with a dense reward designed to stabilize grasping while tracking the commanded spin.
Table~\ref{tab:teacher_reward_shaping} summarizes the reward terms and weights used in simulation.
Episodes terminate on timeout or if the cube drops below a height threshold.

\begin{table}[h]
\centering
\caption{Reward shaping terms used for teacher training in simulation.}
\label{tab:teacher_reward_shaping}
\begin{tabular}{p{0.22\linewidth}p{0.52\linewidth}c}
\toprule
\textbf{Term} & \textbf{Explanation} & \textbf{Weight} \\
\midrule
Object position tracking & Exponential tracking to the origin. & $+10$ \\
Rotation magnitude tracking & Exponential tracking to the command. & $+10$ \\
Rotation direction alignment & Alignment between cube angular velocity and commanded axis. & $+30$ \\
Angular acceleration & Penalty on cube angular acceleration. & $+1$ \\
Reaching object & Encourage finger tips to get close to the object. & $+15$ \\
Pose closure & Encourage the vectors from the finger tips to the object's center of mass to be opposite to each other. & $+2$ \\
\midrule
Joint velocity penalty & $\ell_2$ penalty on joint velocities. & $-10^{-4}$ \\
Action penalty & $\ell_2$ penalty on actions. & $-5\times 10^{-4}$ \\
Action rate penalty & $\ell_2$ penalty on action differences. & $-10^{-2}$ \\
DOF acceleration penalty & $\ell_2$ penalty on joint accelerations. & $-10^{-7}$ \\
\bottomrule
\end{tabular}
\end{table}

\subparagraph{Domain randomization.}
At startup, we scale the robot link friction uniformly from the range $[0.5,1.3]$ and scale robot link masses by $[0.95,1.05]$.
On reset we randomize: actuator stiffness and damping (log-uniform scaling, 0.75--1.5 and 0.3--3.0), small perturbations to joint limits, cube friction (static in $[0.5,1.3]$), cube mass scaling (uniform in $[0.2,1.0]$), cube initial pose/velocity perturbations, and gravity perturbations (Gaussian additive noise with standard deviation up to 0.5\,m/s$^2$ along $z$).

\subparagraph{Teacher PPO training.}
We train the teacher with PPO as implemented in RSL-RL~\citep{schwarke2025rslrl}.
The policy and value networks are MLPs with hidden sizes [512, 512, 256, 128] and ELU activations.
We use observation normalization for both actor and critic, and a learned log standard deviation for the Gaussian policy (initial std 1.0).
The PPO training hyperparameters are summarized in Table~\ref{tab:teacher_ppo_hparams}.

\begin{table}[h]
\centering
\caption{Teacher PPO training hyperparameters.}
\label{tab:teacher_ppo_hparams}
\begin{tabular}{p{0.55\linewidth}p{0.35\linewidth}}
\toprule
\textbf{Hyperparameter} & \textbf{Value} \\
\midrule
Rollout length (steps per env) & 24 \\
Max iterations & 10{,}000 \\
Learning epochs per iteration & 5 \\
Minibatches per epoch & 4 \\
Learning rate & $10^{-4}$ \\
LR schedule & adaptive (target KL) \\
Desired KL & 0.01 \\
Discount factor $\gamma$ & 0.99 \\
GAE parameter $\lambda$ & 0.95 \\
PPO clip $\epsilon$ & 0.2 \\
Entropy coefficient & $10^{-4}$ \\
Value loss coefficient & 1.0 \\
Clipped value loss & enabled \\
Gradient norm clip & 1.0 \\
\bottomrule
\end{tabular}
\end{table}

\paragraph{Policy distillation procedure.}
\label{app:distillation}
We distill a PPO-trained teacher policy into the \algoname model using IsaacLab. We follow the principle as in~\citep{miki2022learning}, with adaptation to chunked actions. During distillation, we applied strong observation noise to the student to improve the robustness. This noise consists of additive Gaussian noise and a random offset that is reset at the beginning of each episode. Noise is applied to both joint positions and velocities as well as to the object state.

We simulate 1{,}024 parallel environments during distillation. At each time step, the teacher policy produces an action that is recorded as a supervisory signal for the student. The executed action is selected stochastically: with probability $p_{\text{teacher}}$, the teacher's action is executed, and with probability $1-p_{\text{teacher}}$, the student's predicted action is executed. A data buffer of student observations and the corresponding teacher actions is populated for 128 time steps. Observation--action pairs are then extracted using a sliding window following the structure illustrated in Fig.~\ref{fig:RTC_setups}.

The student policy is trained using a behavior cloning loss to match the teacher's actions. The probability $p_{\text{teacher}}$ starts at 1.0 and decays by a factor of 0.999 at each iteration to gradually shift the control from the teacher to the student. We initialize the policy by running this distillation process for 200 iterations, corresponding to approximately 20 million gradient steps.

\subsection{Rubber-Duck Pick-and-Place}
\label{app:duck}

\begin{figure}[h]
\centering
\includegraphics[width=0.2\linewidth]{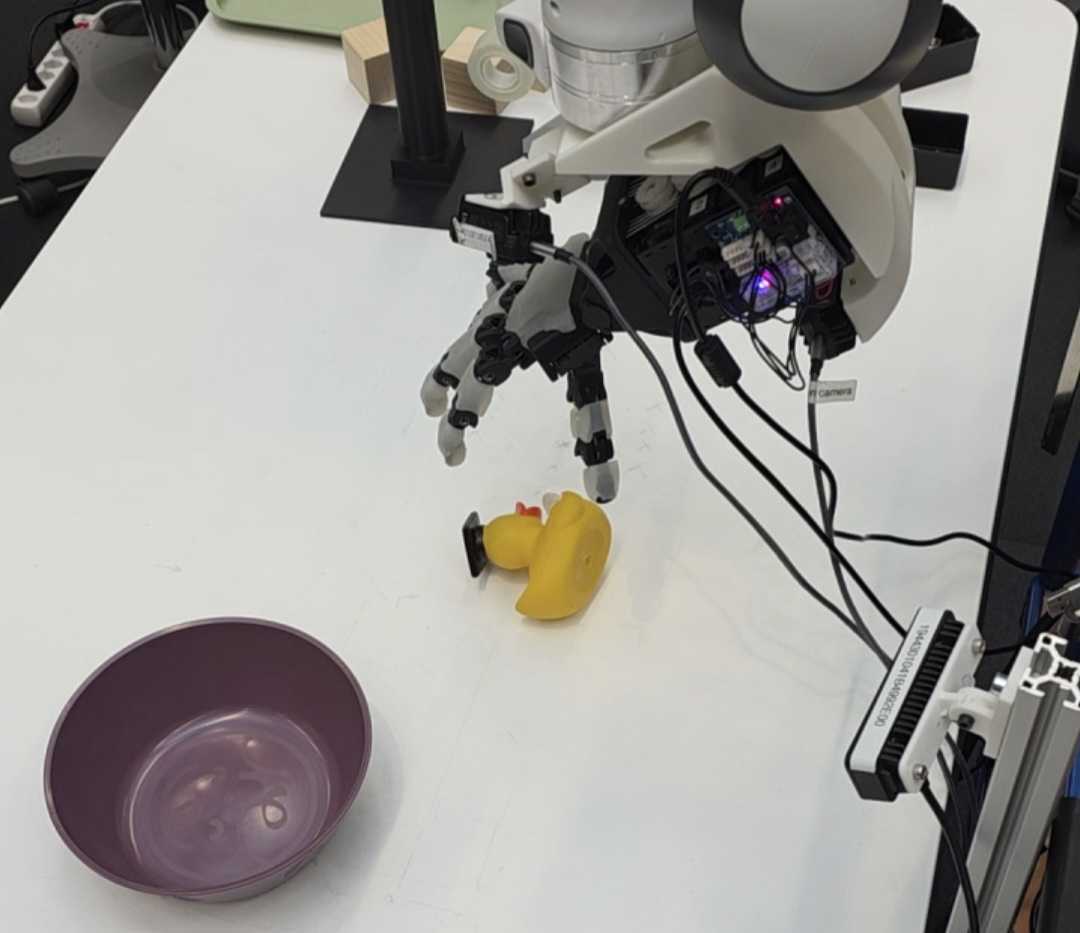}\hfill
\includegraphics[width=0.2\linewidth]{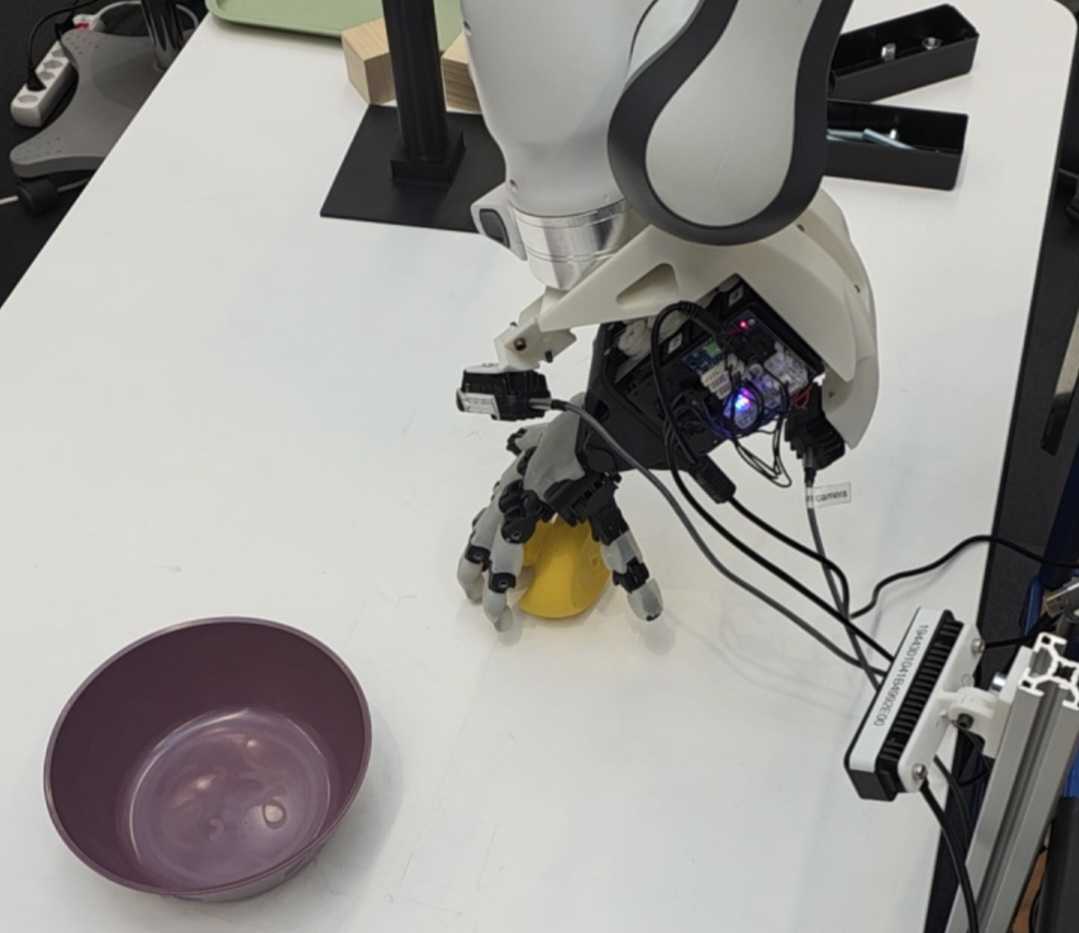}\hfill
\includegraphics[width=0.2\linewidth]{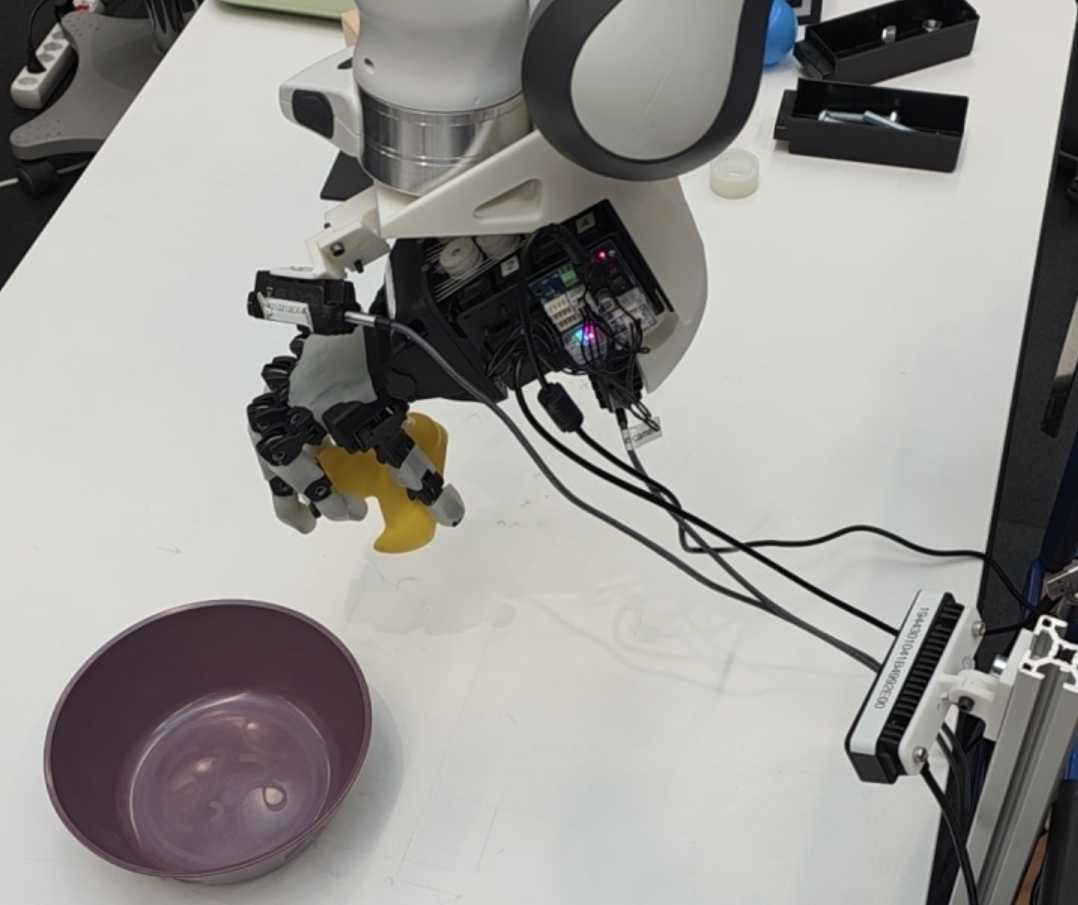}\hfill
\includegraphics[width=0.2\linewidth]{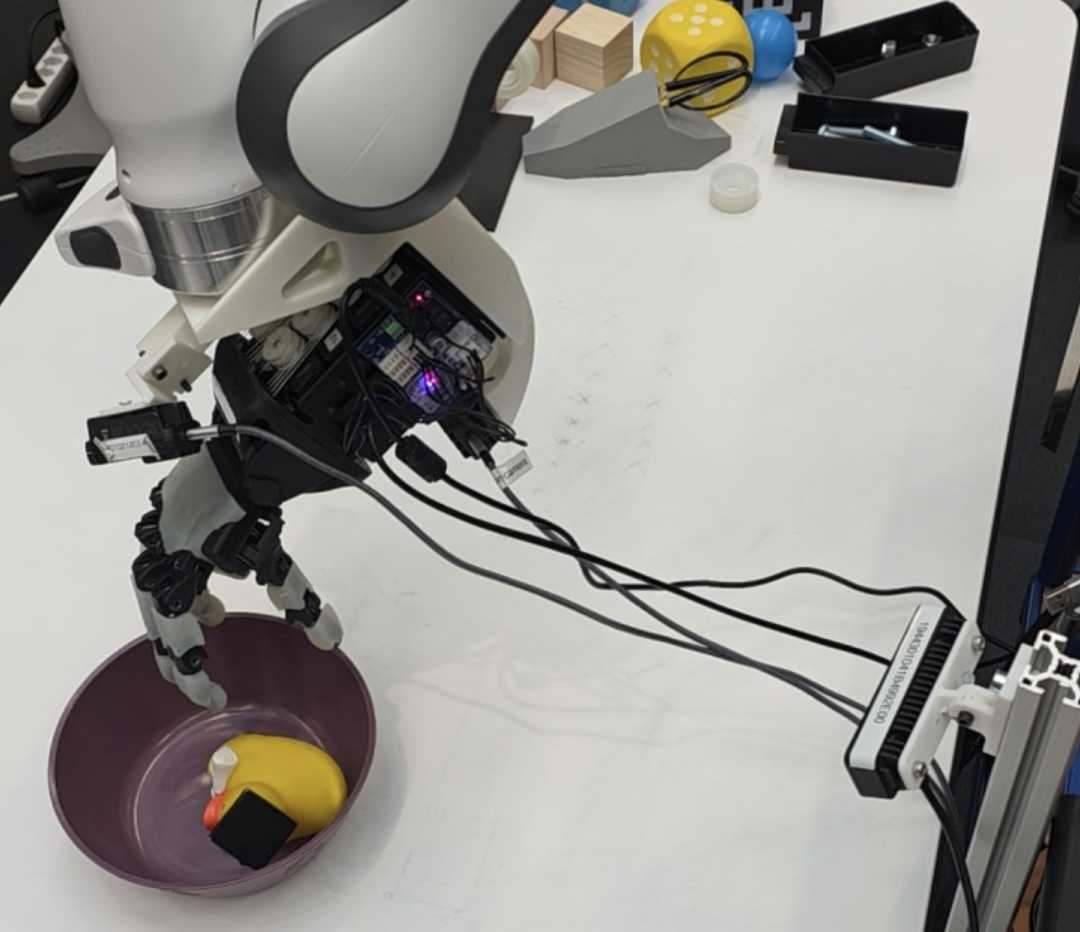}
\caption{The example of rubber-duck pick-and-place task.}
\label{fig:duck_snapshots}
\end{figure}

We use the same Franka + ORCA platform, end-effector control, and image
encoder pipeline as for the scissors task. The duck is initialized in one of
many poses (lying, upside-down, on its side, random in-plane rotations) and
the policy must grasp and place it into a bowl placed near the workspace.
We collect 30 successful teleoperated demonstrations and 15 additional
online rollouts during online RL. Evaluation uses 2 fixed start positions
$\times$ 8 orientations (16 trials). Representative rollout snapshot is
shown in Fig.~\ref{fig:duck_snapshots}.

\section{Network Architecture and Hyperparameters}
\label{app:implementation}

This section describes \algoname's network architecture and the
hyperparameters used across our experiments. We first present the overall
actor--critic architecture (Sec.~\ref{app:architecture_fig}), then detail
the individual network components (Sec.~\ref{app:architectures}),
training hyperparameters (Sec.~\ref{app:hyperparameters}), and baseline
implementations (Sec.~\ref{app:baseline_impl}).

\subsection{Actor--Critic Architecture}
\label{app:architecture_fig}

\begin{figure}[h]
  \centering
  \includegraphics[width=0.85\linewidth]{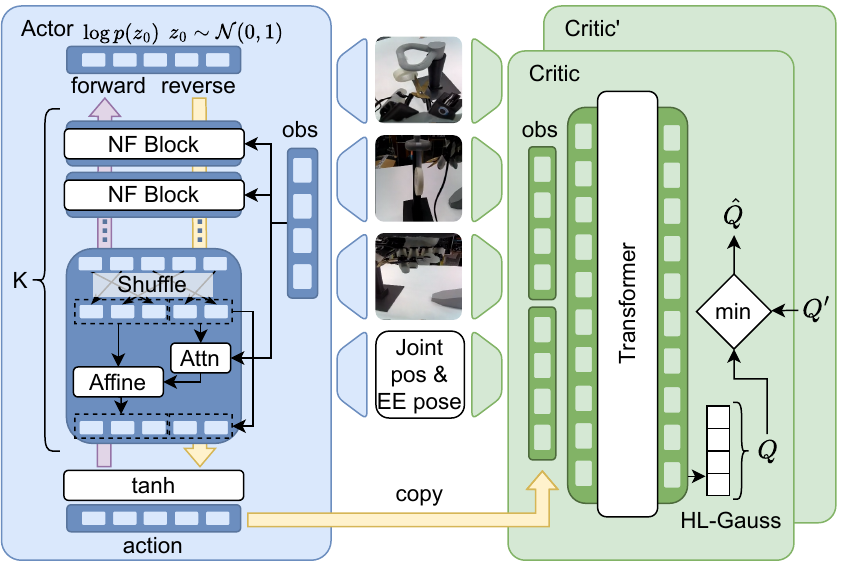}
  \caption{Actor--critic architecture of \algoname. \textbf{Actor:}
  conditional normalizing flow that maps invertibly between action chunks
  and a base Gaussian. Stacked NF blocks apply affine transformations to
  a partitioned subset of tokens conditioned on observations. Forward pass
  yields tractable log-likelihoods used for behavior-cloning
  supervision~\eqref{eq:loss_il_nf}; inverse pass is fully differentiable
  and used for critic-guided updates. \textbf{Critic:} transformer
  Q-network that scores action chunks. Q-values are parameterized as an
  HL-Gaussian categorical
  distribution~\citep{farebrother2024stop} for regression stability, with
  minimum over two critics to mitigate overestimation.}
  \label{fig:network}
\end{figure}

The actor is a conditional normalizing flow that maps action chunks
invertibly to a base Gaussian, enabling both efficient sampling and exact
log-likelihood evaluation. The critic is a transformer Q-network that
scores entire action chunks rather than individual actions, aligning
value estimation with the chunked control interface used at deployment.
A schematic of the two networks and how they share observation tokens is
shown in Fig.~\ref{fig:network}.

\subsection{Network Components}
\label{app:architectures}

Below we list per-component details for the visual encoder, the
normalizing-flow actor, and the chunked critic, along with parameter
counts used in our experiments.

\paragraph{Visual encoders.}
\begin{itemize}
  \item \textbf{ResNet-18 (RoboMimic).} Two separate ResNet-18 backbones are used for \texttt{img0} and \texttt{img1}. Each encoder follows the standard torchvision ResNet-18 up to the last convolutional stage (no global average pooling and no FC classifier). Activation: ReLU. Normalization: BatchNorm2d. Image normalization uses ImageNet mean/std inside the forward pass. For $224\times 224$ inputs, the output is a $7\times 7$ grid of 512-dim tokens (49 tokens).
  \item \textbf{DINOv2 ViT-L/14 (scissors).} The encoder is \texttt{dinov2-vit-l} from \texttt{timm} (ViT-L/14 with 4 register tokens). Images are resized to $224\times 224$ with \texttt{resize-naive}. The backbone outputs patch tokens ($16\times 16$ grid, 256 patches) taken from the penultimate block; a linear projector maps 1024-dim patch features to 512-dim tokens. Activation: GELU. Normalization: LayerNorm. Fine-tuning mode: \texttt{norm} (only LayerNorm parameters are trainable). A single backbone is shared across the three camera views via \texttt{backbone\_instance\_name}.
  \item \textbf{No visual encoder (ORCA in-hand).} Only low-dimensional proprioceptive and command inputs are used.
\end{itemize}

\paragraph{Normalizing-flow actor.}
\begin{itemize}
  \item \textbf{Model.} RealNVP with 16 affine coupling layers and an initial invertible ArcTanh.
  \item \textbf{Coupling network.} Each coupling layer uses a causal TransformerBlock with hidden dim 256, 8 heads, and \texttt{block\_depth} 1. The block consists of FlowBlocks with self-attention, cross-attention, RMSNorm, RoPE positional embeddings, and SwiGLU MLP (SiLU gating). Final scale/shift heads use GELU.
\end{itemize}

\paragraph{Critic network (Q-chunking).}
\begin{itemize}
  \item \textbf{Architecture.} Each critic is a transformer encoder (\texttt{d\_model}=512, \texttt{nhead}=8, 3 encoder layers, FF dim 256, activation GELU, dropout 0.1) operating over observation tokens plus an appended action token.
  \item \textbf{Action token.} Actions are projected via Linear + ELU to 512-dim, with a learned action positional bias.
  \item \textbf{Output.} Distributional Q with 101 bins.
  \item \textbf{Ensemble.} Two critics are used in all NF-Q-chunking configurations.
  \item \textbf{Normalization.} LayerNorm in each TransformerEncoder layer.
\end{itemize}

\paragraph{Parameter counts.}
\begin{itemize}
  \item \textbf{ResNet-18 (per encoder).} 11{,}176{,}512 parameters (no final FC).
  \item \textbf{DINOv2 ViT-L/14 (per backbone + projector).} 303{,}711{,}744 parameters.
  \item \textbf{Flow actor (per policy).} 16 coupling layers; per-layer parameters are
    $3{,}283{,}904 + 1{,}282 \times \text{action\_dim}$, giving total
    $16 \times (3{,}283{,}904 + 1{,}282 \times \text{action\_dim})$.
  \item \textbf{Critic (per critic).} $3{,}999{,}077 + 512 \times (\text{action\_dim} + T_a + 1)$.
\end{itemize}

\subsection{Hyperparameters}
\label{app:hyperparameters}

We apply dropout for $\pi_\theta$ regularization, using a rate of 0.5 during policy
initialization (reduced to 0.2 for real-world experiments to accelerate
convergence under limited compute).
During reinforcement learning, the dropout rate is reduced to 0.1, as higher
values were found to degrade offline RL performance, while a small amount of
dropout remains beneficial~\citep{hu2023imitation,tarasov2024role}.

The number of inverse samples used for action selection is set to 128 in
simulation, 64 in real-world inference after policy initialization, and 24
during RL fine-tuning due to GPU memory constraints on the real robot system.

For the scissors task, the batch size is 256 for imitation learning and 48 for all RL stages. For the cube rotation task, the batch size is 1024 for distillation and 512 for the RL stage.

We use a high discount factor $\gamma=0.997$, which is well suited for the
sparse-reward, long-horizon manipulation tasks considered in this work.
Target networks are updated using Polyak averaging with $\tau=0.05$. During the online RL stage of the simulated tasks, offline and online data are mixed with $\rho=0.5$. In the real-world tasks, during the online reinforcement learning stage, training batches are constructed from a mixture of offline and online data sampled in proportion to their sizes, with offline data assigned a mixing weight of 0.5.

All models are trained using the AdamW optimizer~\citep{loshchilov2017decoupled}
with weight decay set to $10^{-4}$ throughout.
The learning rate is $10^{-4}$ during policy initialization and increased to
$2\times 10^{-4}$ during reinforcement learning to accelerate adaptation. We use default Adam values $\beta_1=0.9$, $\beta_2=0.999$, $\epsilon=10^{-8}$.

For the scissors task, we first train imitation learning policy with 200{,}000 update steps, then the critic warm-up stage consists of 5{,}000 gradient updates, full offline RL of 1{,}000 steps, and online RL of 4{,}000 steps.

For the cube rotation task, we first distill the policy using 20\,M updates, then 3{,}000 steps of critic warm-up, 1{,}000 steps of full offline RL, and 3{,}000 steps of online RL.

For simulation tasks, we train imitation learning policies for 30{,}000 steps.

\subsection{Baseline Implementation Details}
\label{app:baseline_impl}

We compare \algoname against strong baseline methods to evaluate the expressiveness and accuracy of our approach in the imitation learning setting. To ensure a fair comparison, the dataset, optimizer settings, data augmentation strategies, and image encoders are kept identical across all methods.

\paragraph{Action Chunking Transformer.}
We implement the Action Chunking Transformer (ACT)~\citep{zhao2023ACT} using a Transformer-based architecture. A chunk size of 20 is used, as larger chunk sizes are typically required to achieve strong performance with ACT~\citep{zhao2023ACT}. The style variable has a dimensionality of 32, and the KL divergence loss is weighted by 10.

Conditioned only on low-dimensional state inputs, the action encoder is a transformer with a model dimension of 512, 8 attention heads, 3 layers, a feedforward dimension of 2048, and a dropout rate of 0.1. The action decoder uses the same configuration and is conditioned on both low-dimensional inputs and visual observations. Causal masking is applied to both the encoder and decoder.

Temporal aggregation is used to ensure smooth and accurate action execution, with an exponential weighting parameter of $k = 0.01$.

\paragraph{Flow matching.}
The flow-matching baseline is implemented with the DiT block policy backbone~\citep{dasari2025ingredients}. The model has a dimension of 512, 8 attention heads, 4 DiT blocks, a feedforward dimension of 2048, a dropout rate of 0.1, and uses 8 flow steps. Causal masking is applied throughout the model.
Both training and inference follow the procedure described in~\cite{black2024pi_0}.
Real-time chunking~\cite{black2025training} is applied, with the same configuration as our \algoname, as shown in Fig.~\ref{fig:RTC_setups}.

\section{Additional Experimental Results}
\label{app:additional_results}

\subsection{Simulation Validation on RoboMimic}
\label{app:robomimic_results}

We first validate \algoname in simulation to identify reasonable hyperparameters
before deploying on real hardware. We use the RoboMimic
benchmark~\citep{mandlekar2021matters} with the Lift, Can, and Square
manipulation tasks; for all tasks we use the Mixed Human (MH) datasets,
which mix successful and suboptimal demonstrations and are therefore more
challenging than Expert Human datasets. Policies operate on image
observations at $84\times 84$, encoded with a ResNet-18~\citep{he2015deep}
backbone; success rate is evaluated over 100 rollouts per seed.

\begin{table}[h]
\centering
\caption{Success rates (over 4 seeds, 100 evaluation rollouts each) on
RoboMimic MH datasets. All hyperparameters are tuned on Square and reused
for Lift and Can.}
\label{tab:robomimic_results}
\begin{tabular}{l|cc}
\toprule
\textbf{Environment} & \textbf{Imitation Learning} & \textbf{Offline RL} \\
\midrule
Lift   & $0.79 \pm 0.00$ & $0.91 \pm 0.04$ \\
Can    & $0.96 \pm 0.00$ & $0.96 \pm 0.02$ \\
Square & $0.61 \pm 0.04$ & $0.68 \pm 0.05$ \\
\bottomrule
\end{tabular}
\end{table}

Offline RL improves over IL when the initial policy does not already solve
the task: Lift gains 12 absolute points, Square 7. Can is saturated by IL.
The hyperparameter set transfers across tasks without modification and is
reused in our real-world experiments.

\subsection{Ablation Studies}
\label{app:ablations}

\begin{figure}[h]
    \centering
    \includegraphics[width=\linewidth]{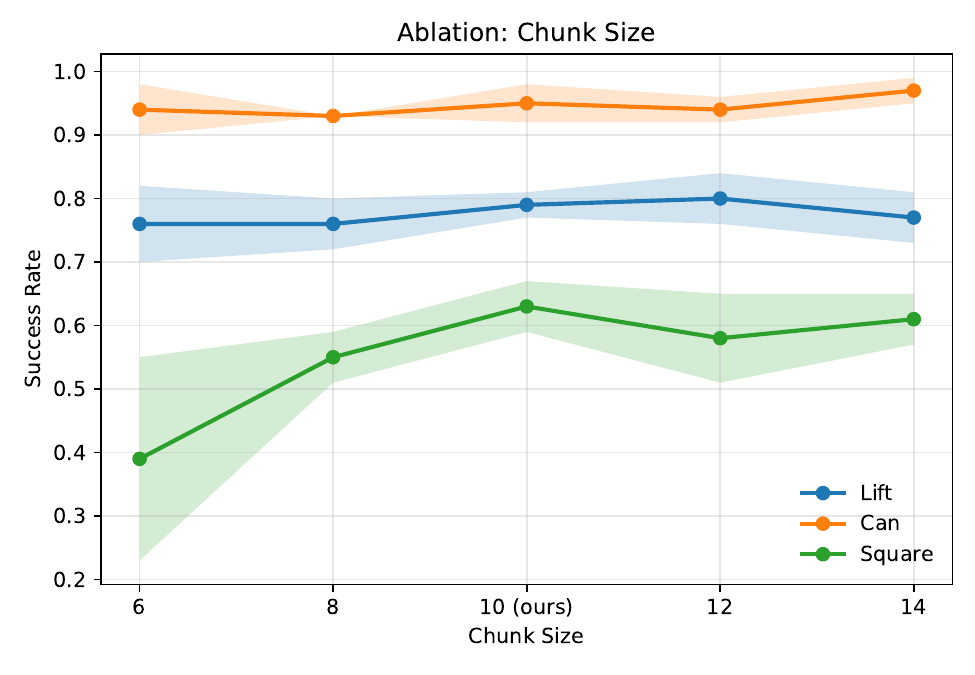}
    \caption{Effect of action chunk length $H$ on imitation learning performance
across RoboMimic Lift, Can, and Square tasks with 4 random seeds.}
    \label{fig:abl_chunk}
\end{figure}

\begin{figure}[h]
    \centering
    \includegraphics[width=\linewidth]{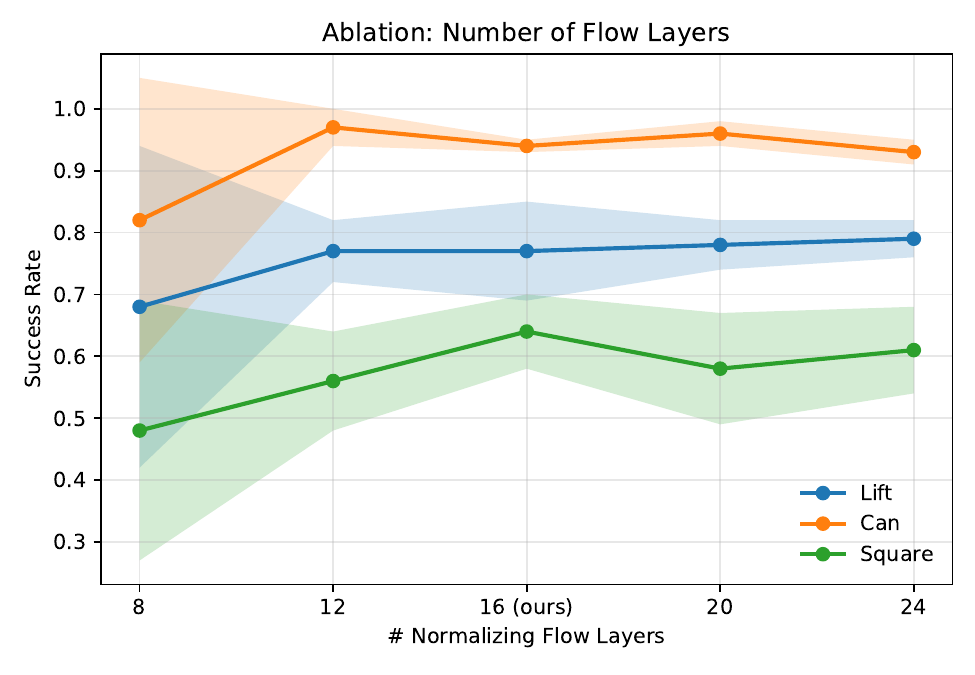}
    \caption{Effect of normalizing-flow depth (number of coupling blocks) on
imitation learning performance across RoboMimic tasks with 4 random seeds.}
    \label{fig:abl_layers}
\end{figure}

\begin{figure}[h]
    \centering
    \includegraphics[width=\linewidth]{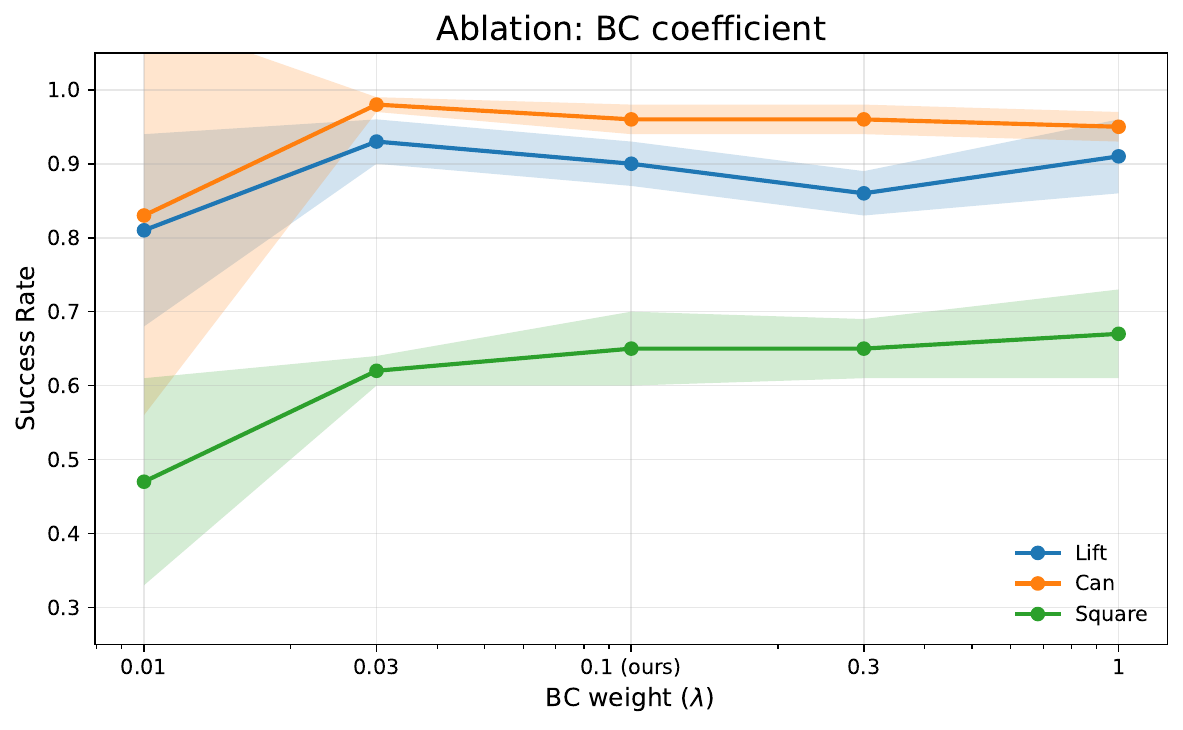}
    \caption{Effect of $\lambda$ on
offline RL performance across RoboMimic tasks with 4 random seeds.}
    \label{fig:abl_bc}
\end{figure}

We conduct ablation studies to analyze the sensitivity of \algoname to key
architectural and algorithmic design choices.
The ablations in this subsection are performed on simulated RoboMimic environments
(Lift, Can, and Square).

\paragraph{Effect of action chunk length.}
Fig.~\ref{fig:abl_chunk} shows the effect of varying the action chunk length
$H \in \{6,8,10,12,14\}$.
Across all environments, intermediate chunk sizes yield the most consistent
performance.
Our choice of $H=10$ performs robustly across Lift ($0.79 \pm 0.02$),
Can ($0.95 \pm 0.03$), and Square ($0.63 \pm 0.04$), balancing temporal
abstraction and controllability.
Shorter chunks ($H=6$) lead to a clear degradation on the Square task
($0.39 \pm 0.16$), while larger chunks do not consistently improve performance
and can slightly reduce stability due to increased open-loop execution.

\paragraph{Effect of normalizing-flow capacity.}
Fig.~\ref{fig:abl_layers} ablates the number of coupling blocks in the
normalizing-flow policy, varying the depth from 8 to 24 layers.
Performance improves with increasing capacity up to a point and then
saturates.
A model with 16 coupling blocks achieves strong and stable performance across
all tasks (Lift: $0.77 \pm 0.08$, Can: $0.94 \pm 0.01$, Square: $0.64 \pm 0.06$),
while shallower models underfit and deeper models provide only marginal gains
at higher computational cost.
Based on this trade-off, we adopt 16 coupling blocks in all experiments.

\paragraph{Effect of BC coefficient.}
Fig.~\ref{fig:abl_bc} studies the impact of the behavior cloning weight $\lambda \in \{0.01, 0.03, 0.1, 0.3, 1.0\}$ on offline RL performance.
We observe that performance is sensitive to this hyperparameter, with very small values leading to unstable learning (especially on Can), while overly large values bias the policy toward imitation and can reduce returns on Lift.
Overall, $\lambda=0.1$ provides a good balance between stabilizing policy optimization and allowing improvement beyond the dataset, and we therefore use $\lambda=0.1$ in all experiments.

\subsection{ACT + Offline RL Negative Result}
\label{app:act_rl}

We attempted to fine-tune an ACT~\citep{zhao2023learning} backbone with our
chunked-critic and TD3+BC-style MSE behavior-cloning regularizer. The actor
diverged almost immediately after warm-up: success rates on grasping and
cutting both collapsed to 0. We attribute this to the high sensitivity of
the MSE regularizer weight $\lambda_{\text{MSE}}$ in the presence of
multimodal data -- the same setting that motivates the likelihood-based
regularizer used by \algoname. Within the available time and compute we
could not find a $\lambda_{\text{MSE}}$ value at which training was stable.

\subsection{IL Ablations: Effect of Additional Data}
\label{app:il_ablations}

To verify that \algoname's gains come from critic-guided improvement
rather than extra data, we evaluate two IL-only ablations on the scissors
task. All entries in Table~\ref{tab:il_ablations} are evaluated under a
reduced 10-configuration protocol, because re-running these IL ablations
with the full 25-configuration evaluation grid was not feasible within
our time budget. As a result, the \algoname (Offline only) and \algoname
(Full) rows in this table are under the 10-configuration protocol and so
do not directly match the aggregated 25-trial numbers reported for
\algoname in the main paper; they are repeated here only for context.
Conclusions are unchanged across protocols.

In Table~\ref{tab:il_ablations}, the $N$ column reports the number of
trajectories used to train each IL policy, with the $+$ superscript
denoting successful trajectories. For \algoname rows, we report
$N_{\text{total}}\,(N^+)$: total available trajectories and the
successful subset (only the successful subset is used by the IL
initialization; the full set is used by offline RL).

\begin{table}[h]
\centering
\caption{Scissors-task IL ablations under the original 10-configuration
evaluation protocol. $+$ superscript denotes successful trajectories used
for IL.}
\label{tab:il_ablations}
\begin{tabular}{l|ccc}
\toprule
\textbf{Policy} & $N$ \textbf{trajectories} & \textbf{Grasping} & \textbf{Cutting} \\
\midrule
ACT IL                    & $71^+$       & 0.5 & 0.0 \\
Flow-Matching IL          & $71^+$       & 0.5 & 0.1 \\
\midrule
NF IL (original data)     & $71^+$       & 0.5 & 0.1 \\
NF IL + online RL data    & $83^+$       & 0.1 & 0.0 \\
NF IL + more teleoperation& $97^+$       & 0.6 & 0.0 \\
\midrule
\algoname (Offline only)  & $121$ ($71^+$) & \textbf{0.8} & 0.0 \\
\algoname (Full)          & $181$ ($83^+$) & 0.7 & \textbf{0.7} \\
\bottomrule
\end{tabular}
\end{table}

Two findings are robust under this protocol. First, simply augmenting
the original demonstrations with 26 additional teleoperated trajectories
(NF IL + more teleoperation, $N{=}97^+$) yields only marginal gains
($0.6/0.0$ grasp/cut) and does not enable cutting at all. Second,
naively pooling on-policy trajectories collected during online RL with
the offline IL dataset (NF IL + online RL data, $N{=}83^+$)
\emph{hurts} performance ($0.1/0.0$), likely due to distribution
mismatch between expert teleoperation and on-policy data. In contrast,
\algoname leverages both the failure trajectories and the on-policy
data through critic-guided RL updates, achieving substantially higher
performance with the same data budget. These conclusions are
consistent with the main-paper results under the expanded
15-configuration grid (Table~\ref{tab:baselines},
Sec.~\ref{sec:results_scissors}).

\subsection{Failure Mode Analysis}
\label{app:failures}

The primary failure modes observed in the scissors task include grasp failure, scissor dropping, and task timeout. Achieving a stable grasp is particularly challenging due to the absence of tactile sensing and frequent occlusion of the index finger by the thumb in wrist-mounted camera views. In both teleoperation and autonomous execution, the fingers must be inserted deeply into the scissor handles to achieve a secure grasp, which further increases sensitivity to perception and control errors. Task timeouts are typically caused by accumulated positioning errors while navigating toward the tape. Additionally, the tape's semi-transparent appearance makes it difficult for the front-facing camera to accurately estimate the relative depth between the scissors and the tape, leading to imprecise cutting motions.

For the cube reorientation task, the dominant failure modes are cube dropping and lack of motion. Cube localization is highly noisy due to frequent occlusion by the fingers, resulting in a significant sim-to-real gap that degrades policy performance.

\section{Computational Requirements and Software Details}
\label{app:computation}

Training of real-world policies was performed on NVIDIA H200 and A100 GPUs (single-GPU jobs),
while policy inference and cube-task distillation were run on an NVIDIA RTX
4090.
Simulation experiments were carried out on NVIDIA TITAN RTX GPUs.
GPUs were utilized at full capacity to accelerate training and inference;
however, all stages can be executed on less powerful hardware by reducing
batch sizes or the number of sampled candidate actions during policy
evaluation.

For the scissors task, imitation learning pretraining required approximately
35 hours, followed by 21 hours of offline reinforcement learning.
For the cube task, simulation-to-real distillation required approximately
4 days on a consumer-grade GPU, while the subsequent real-world offline RL
stage completed in about 2 hours.

At inference time, the policy executes one action chunk of length $H=10$ in
approximately 0.24 seconds for the most compute-intensive scissors task on an
NVIDIA RTX 3090 Ti.
For the cube task, inference latency is below 0.1 seconds per action chunk.
These runtimes are compatible with real-time control in all evaluated tasks.

During on-policy training of the scissors task, gradient updates are performed on an NVIDIA H200 GPU, while policy inference and data collection run on a separate desktop equipped with an NVIDIA RTX 3090. The training and inference processes operate asynchronously and coordinate through file transfers and human commands: model checkpoints and experience buffers are stored on disk and transferred between machines.
The inference process loads the latest available checkpoint for policy execution, while newly collected experience is saved to disk and made available to the training process. The model is implemented in PyTorch, and experience buffers are stored on disk using Zarr.

\section{Bimanual Tube and Ball Pickups}
\label{app:biman}

To probe \algoname under a \emph{bimanual} regime that stresses
coordinated two-arm control, deformable-object grasping, and a much
larger action space than any of our single-arm tasks, we introduce a
real-world bimanual pickup task. The robot is required to (i) pick up
a thin-walled plastic bottle (used here as a tube for the ball) from a
horizontal pose on the table with the left arm and hand, and (ii) pick
up a tennis ball from the table with the right arm and hand. The task
is challenging because the plastic bottle is hard to grasp robustly --
gripping too lightly causes it to slip out of the hand, while gripping
too hard deforms it and, combined with an incorrect finger
configuration, also causes slippage; the tennis ball moves easily on
contact, requiring careful approach and finger placement; and the
combined bimanual action space is $48$ DoF (two 7-DoF end-effector
poses plus two 17-DoF ORCA hand command vectors), substantially
larger than any single-arm task in this paper and considerably harder
to learn from limited real-world data.

\subsection{Experimental Setup}
\label{app:biman_setup}

\begin{figure}[h]
     \centering
     \includegraphics[width=0.7\linewidth]{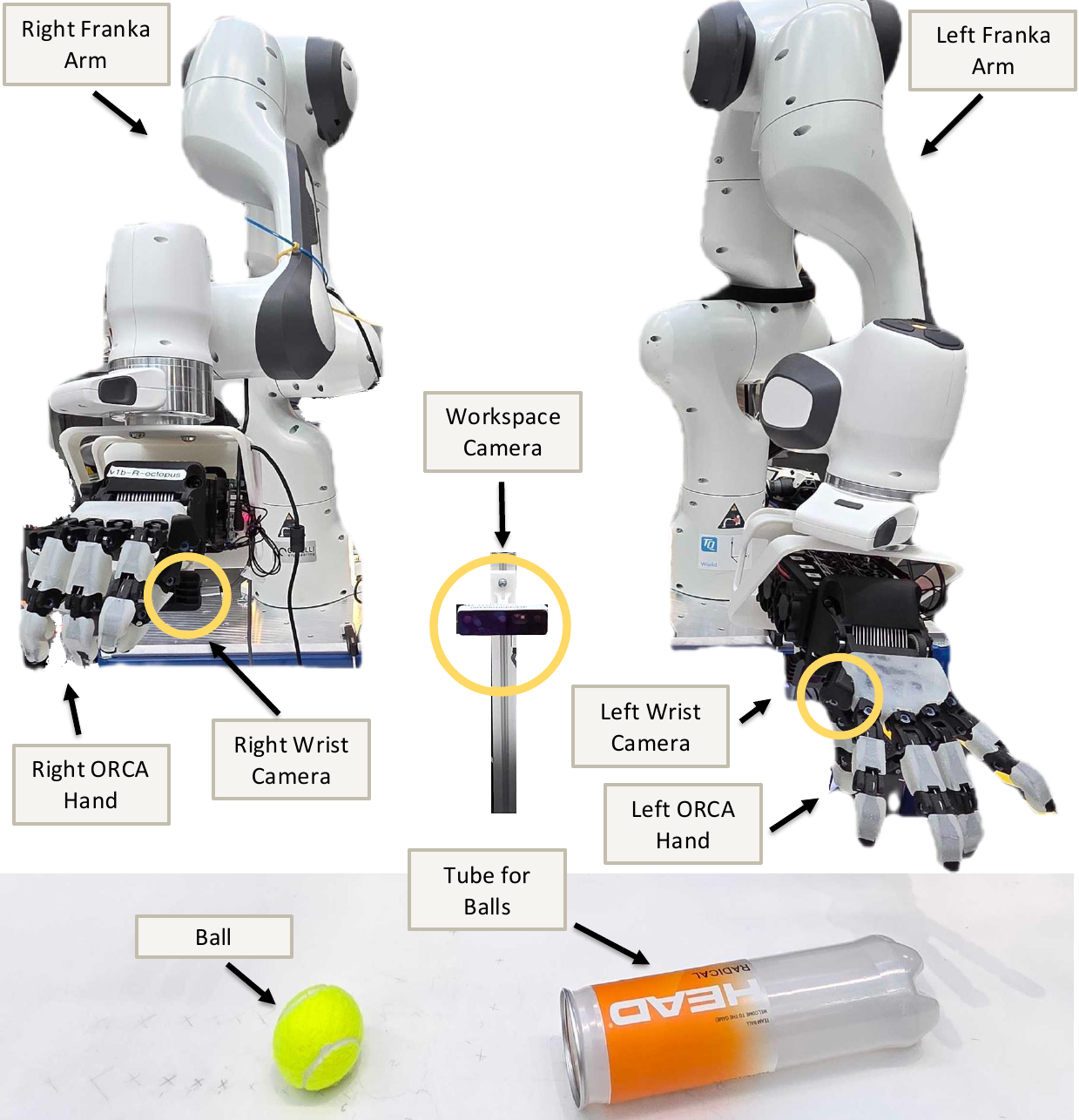}
     \caption{Real-world setup for the bimanual tube and ball pickup
     task. Two 7-DoF Franka Emika Panda arms are each equipped with an
     ORCA dexterous hand. A single workspace camera observes the
     tabletop scene from in front of the workspace, and each hand
     additionally carries one wrist-mounted RGB camera providing a
     close-up view from below the corresponding palm. The tube starts
     (a thin-walled plastic bottle) in a horizontal pose on the table
     within reach of the left hand; the tennis ball is placed within
     reach of the right hand.}
     \label{fig:biman_setup}
\end{figure}

\paragraph{Hardware setup.}
The setup uses two 7-DoF Franka Emika Panda arms mounted on a shared
table base, each equipped with an ORCA
hand~\citep{christoph2025orcaopensourcereliablecosteffective} via a
custom 3D-printed mount, giving a combined action space of
$2\times 24 = 48$ DoF (end-effector pose plus per-hand joint commands
for each arm). Visual observation consists of three RGB streams: a
workspace camera mounted in front of the table providing a
third-person view of the scene, and one wrist-mounted camera below
each hand giving a close-up of the grasped object. All streams are
cropped and resized to $224\times 224$ and encoded with the same
frozen DINOv2 backbone~\citep{oquab2023dinov2} used for the scissors
task. The policy runs at $10$\,Hz; each Franka arm is controlled by a
separate low-level NUC at $1$\,kHz, and hand commands are streamed at
$40$\,Hz. Fig.~\ref{fig:biman_setup} illustrates the setup.

\paragraph{Teleoperation and data collection.}
Demonstrations are collected with a single operator teleoperating the two
arms simultaneously using Rokoko Smart Gloves and Rokoko Coil Pro, with the same energy-based hand
retargeting~\citep{sivakumar2022retargeter} as for the scissors task.
Across the dataset we use a mixture of teleoperation data collected by
\emph{three different people}, which makes the resulting
demonstration distribution highly multimodal: approach trajectories,
grasp shapes, and bimanual timing patterns differ substantially across
operators, putting additional pressure on the expressiveness of the
policy class.
We collect $195$ teleoperated demonstrations, of which $123$ are fully
successful (both pickups completed in the same episode); the remaining
$72$ are partial or failed attempts. The dominant teleoperation
failures are tube slip out of the left hand and ball roll-away when
the right hand contacts it; failures are not discarded and are reused
as offline-RL data. Demonstrations are resampled at $10$\,Hz.

\paragraph{Reward labeling.}
Rewards are sparse and manually annotated: $+1$ is given independently
for each successfully completed sub-task, i.e.\ (i) the tube is grasped
and lifted off the table by the left arm and hand, and (ii) the tennis
ball is grasped and lifted off the table by the right arm and hand. We
do not impose a hold-time threshold: a bad grasp causes the object to
drop almost immediately, so a successful lift already implies a stable
grasp in practice. The two sub-tasks are scored independently, so the
maximum per-episode return is $2$.

\paragraph{Action chunking and real-time inference.}
We reuse the real-time chunking
configuration~\citep{black2025real} from the scissors task ($H{=}10$,
$3$ prefix actions, $1$ observation step). The action chunks for the
two hands are \emph{not} predicted independently; instead, at each
decision step the policy emits a single joint action chunk whose
per-step entries are the concatenation
$[a^{\text{left}}_t,\,a^{\text{right}}_t] \in \mathbb{R}^{48}$ of the
left- and right-arm actions, so the actor and critic both reason
about the two arms jointly.

\paragraph{Pipeline and budgets.}
We run the full
IL\,$\to$\,offline-RL\,$\to$\,online-RL pipeline. IL is trained for
$14{,}000$ gradient steps on the $123$ successful demonstrations;
offline RL adds $6{,}000$ steps using all $195$ trajectories with the
sub-task rewards above; online RL then proceeds in iterations of
$1{,}000$ gradient updates each, with new rollouts collected between
iterations and added to the replay buffer (Sec.~\ref{sec:method_online}).
Across the four online iterations reported here, the total number of
trajectories grows from $195$ (post-offline) to $250$, $329$, $477$,
and $566$, corresponding to roughly $55$, $79$, $148$, and $89$ new
on-policy rollouts per iteration. All other hyperparameters (chunk
length, NF depth, $\lambda$, dropout, learning rates) are taken
directly from the scissors configuration without modification
(Appendix~\ref{app:hyperparameters}).

\paragraph{Evaluation protocol.}
Performance is evaluated using the two per-sub-task success rates
(tube pickup, ball pickup) on a fixed $4\times 4$ grid of starting
configurations -- $4$ predefined tube positions $\times$ $4$
predefined ball positions, for a total of $16$ deterministic
evaluation rollouts per checkpoint. We do \emph{not} randomize the
starting positions across evaluations, so all stages are compared on
exactly the same $16$ initial conditions. Episodes terminate when
both pickups are completed, when the tube or ball is dropped after a
grasp, when the robot reaches an unsafe configuration, or after a
$30$\,s timeout.

\subsection{Results}
\label{app:biman_results}

\begin{figure}[h]
     \centering
     \includegraphics[width=0.95\linewidth]{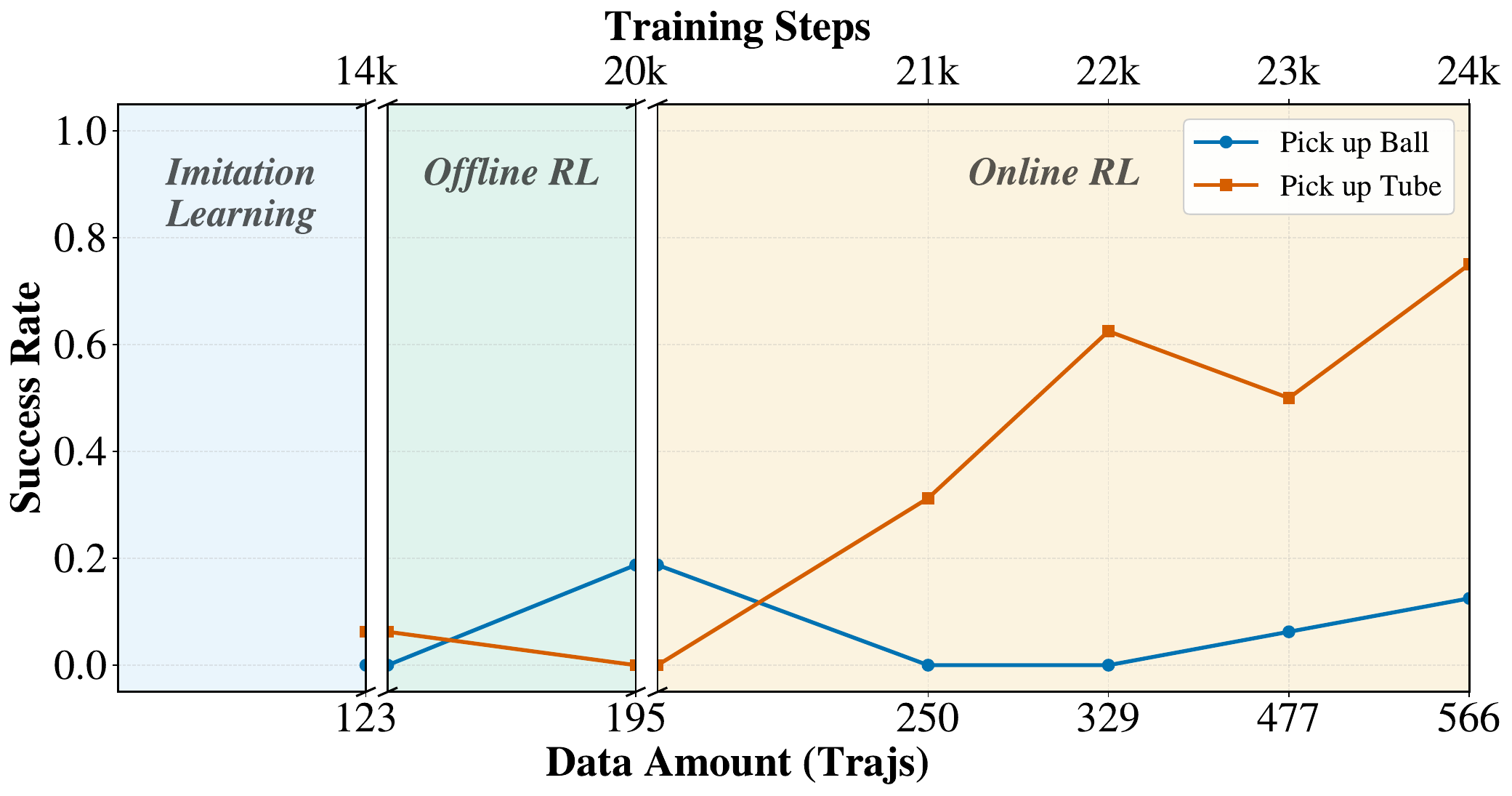}
     \caption{Bimanual tube and ball pickup: per-sub-task success rate
     (over $16$ rollouts) versus training stage. \emph{Imitation
     Learning} is trained for $14{,}000$ steps on the $123$ successful
     demonstrations (data point at $123$ trajectories). \emph{Offline
     RL} adds $6{,}000$ steps using all $195$ teleoperated
     trajectories. \emph{Online RL} then performs four iterations of
     $1{,}000$ updates each, growing the replay buffer to $250$, $329$,
     $477$, and $566$ trajectories. Tube success climbs from $1/16$
     (IL) to $12/16$ ($75\%$) after the fourth online iteration, while
     ball success remains low across all stages, peaking at $3/16$
     ($18.75\%$) after offline RL.}
     \label{fig:biman_curve}
\end{figure}

Fig.~\ref{fig:biman_curve} and Table~\ref{tab:biman_results} summarize
per-sub-task success rates on the bimanual pickup task across the IL,
offline-RL, and four online-RL checkpoints.

\begin{table}[h]
\centering
\caption{Bimanual tube and ball pickup task: per-sub-task success
rates (each over $16$ rollouts) at each training stage. Online RL
iterations report the total trajectory count (teleop + collected
online rollouts) after that iteration.}
\label{tab:biman_results}
\begin{tabular}{l|cc|c}
\toprule
\textbf{Stage} & \textbf{Tube pickup} & \textbf{Ball pickup} & \textbf{Trajectories} \\
\midrule
NF IL (14k)              & 0.06 ($1/16$)   & 0.00 ($0/16$)   & 123 \\
\algoname (Offline, +6k) & 0.00 ($0/16$)   & \textbf{0.19} ($3/16$) & 195 \\
\algoname (Online +1k)   & 0.31 ($5/16$)   & 0.00 ($0/16$)   & 250 \\
\algoname (Online +2k)   & 0.63 ($10/16$)  & 0.00 ($0/16$)   & 329 \\
\algoname (Online +3k)   & 0.50 ($8/16$)   & 0.06 ($1/16$)   & 477 \\
\algoname (Online +4k)   & \textbf{0.75} ($12/16$) & 0.13 ($2/16$) & 566 \\
\bottomrule
\end{tabular}
\end{table}

The two sub-tasks show qualitatively different learning dynamics.
\textbf{Tube pickup} improves substantially with online RL: from
$0.06$ at IL and $0.00$ after offline RL alone, online RL drives the
success rate up to $0.31$, $0.63$, $0.50$, and finally $0.75$ across
the four iterations. We attribute this to critic-guided actor updates
discovering a tube grasp that is firm enough to prevent slip yet not
so firm that the plastic-bottle tube deforms out of the hand -- a
precision
trade-off the IL data alone does not resolve. \textbf{Ball pickup},
in contrast, remains low across all stages: NF IL fails on every
trial, offline RL briefly lifts success to $0.19$ by leveraging the
$72$ partially-failed demonstrations, and online RL recovers only
$0.06$--$0.13$ in the last two iterations. We attribute the low ball
success to (i) the tennis ball rolling away under light contact, which
the policy is rarely able to recover from within a single chunked
attempt; (ii) the very high-dimensional ($48$ DoF) action space, which
makes the limited number of successful ball pickups in the data set
insufficient signal for the critic to single out the correct approach;
and (iii) the fact that we deliberately kept the NF policy capacity
(depth, hidden width) identical to the single-arm configuration even
though the action space has doubled and become considerably more
complex -- the actor may simply not be expressive enough to model the
bimanual action distribution faithfully. We plan to investigate
scaling the NF actor with the action dimensionality in future work.
Crucially, both sub-tasks show \emph{positive} progress through the
pipeline: tube pickup improves by an order of magnitude and ball
pickup recovers a non-trivial signal after offline RL and again at the
end of online RL. We therefore expect that with a larger online-RL
budget -- additional iterations and a correspondingly larger replay
buffer -- both pickups would reach a decent success rate. The cost of
this extra budget is much lower than it would be for teleoperation:
on-policy rollouts during online RL are collected substantially faster
than human teleoperation (no human-in-the-loop, no retargeting
debugging) and, importantly, a large fraction of the collected
trajectories during early online iterations are \emph{quick} failures
(the policy drops the object almost immediately and the episode
terminates well before the $30$\,s timeout), so each iteration
contributes many failure-mode-rich rollouts to the replay buffer at
low real-world time cost. These observations indicate that the
remaining gap on the bimanual task is primarily a budget and model
capacity issue rather than a fundamental limitation of the method.

%% file: references.bib
@article{Chi2023DiffusionPolicy,
  author       = {Cheng Chi and
                  Siyuan Feng and
                  Yilun Du and
                  Zhenjia Xu and
                  Eric Cousineau and
                  Benjamin Burchfiel and
                  Shuran Song},
  title        = {Diffusion Policy: Visuomotor Policy Learning via Action Diffusion},
  journal      = {CoRR},
  volume       = {abs/2303.04137},
  year         = {2023},
  url          = {https://doi.org/10.48550/arXiv.2303.04137},
  doi          = {10.48550/ARXIV.2303.04137},
  eprinttype    = {arXiv},
  eprint       = {2303.04137},
  bibsource    = {dblp computer science bibliography, https://dblp.org}
}

@article{ze20243DDiffusion,
  author       = {Yanjie Ze and
                  Gu Zhang and
                  Kangning Zhang and
                  Chenyuan Hu and
                  Muhan Wang and
                  Huazhe Xu},
  title        = {3D Diffusion Policy},
  journal      = {CoRR},
  volume       = {abs/2403.03954},
  year         = {2024},
  url          = {https://doi.org/10.48550/arXiv.2403.03954},
  doi          = {10.48550/ARXIV.2403.03954},
  eprinttype    = {arXiv},
  eprint       = {2403.03954},
  bibsource    = {dblp computer science bibliography, https://dblp.org}
}

@inproceedings{Ren2025DPPO,
  author       = {Allen Z. Ren and
                  Justin Lidard and
                  Anthony Simeonov and
                  Lars Lien Ankile and
                  Pulkit Agrawal and
                  Anirudha Majumdar and
                  Benjamin Burchfiel and
                  Hongkai Dai and
                  Max Simchowitz},
  title        = {Diffusion Policy Policy Optimization},
  booktitle    = {The Thirteenth International Conference on Learning Representations,
                  {ICLR} 2025, Singapore, April 24-28, 2025},
  publisher    = {OpenReview.net},
  year         = {2025},
  url          = {https://openreview.net/forum?id=mEpqHvbD2h},
  bibsource    = {dblp computer science bibliography, https://dblp.org}
}

@article{McAllister2025FPO,
  author       = {David McAllister and
                  Songwei Ge and
                  Brent Yi and
                  Chung Min Kim and
                  Ethan Weber and
                  Hongsuk Choi and
                  Haiwen Feng and
                  Angjoo Kanazawa},
  title        = {Flow Matching Policy Gradients},
  journal      = {CoRR},
  volume       = {abs/2507.21053},
  year         = {2025},
  url          = {https://doi.org/10.48550/arXiv.2507.21053},
  doi          = {10.48550/ARXIV.2507.21053},
  eprinttype    = {arXiv},
  eprint       = {2507.21053},
  bibsource    = {dblp computer science bibliography, https://dblp.org}
}

@article{Li2025QChunking,
  author       = {Qiyang Li and
                  Zhiyuan Zhou and
                  Sergey Levine},
  title        = {Reinforcement Learning with Action Chunking},
  journal      = {CoRR},
  volume       = {abs/2507.07969},
  year         = {2025},
  url          = {https://doi.org/10.48550/arXiv.2507.07969},
  doi          = {10.48550/ARXIV.2507.07969},
  eprinttype    = {arXiv},
  eprint       = {2507.07969},
  bibsource    = {dblp computer science bibliography, https://dblp.org}
}

@inproceedings{
anonymous2025policyflow,
title={PolicyFlow: Policy Optimization with Continuous Normalizing Flow in Reinforcement Learning},
author={Anonymous},
booktitle={Submitted to The Fourteenth International Conference on Learning Representations},
year={2025},
url={https://openreview.net/forum?id=YETCQLcKtn},
note={under review}
}

@inproceedings{rezende2015variational,
  title={Variational inference with normalizing flows},
  author={Rezende, Danilo and Mohamed, Shakir},
  booktitle={International conference on machine learning},
  pages={1530--1538},
  year={2015},
  organization={PMLR}
}

@article{Ghugare2025NFCapable,
  author       = {Raj Ghugare and
                  Benjamin Eysenbach},
  title        = {Normalizing Flows are Capable Models for {RL}},
  journal      = {CoRR},
  volume       = {abs/2505.23527},
  year         = {2025},
  url          = {https://doi.org/10.48550/arXiv.2505.23527},
  doi          = {10.48550/ARXIV.2505.23527},
  eprinttype    = {arXiv},
  eprint       = {2505.23527},
  bibsource    = {dblp computer science bibliography, https://dblp.org}
}

@article{Lind2025NFCapableVisuomotor,
  author       = {Simon Kristoffersson Lind and
                  Jialong Li and
                  Maj Stenmark and
                  Volker Kr{\"{u}}ger},
  title        = {Normalizing Flows are Capable Visuomotor Policy Learning Models},
  journal      = {CoRR},
  volume       = {abs/2509.21073},
  year         = {2025},
  url          = {https://doi.org/10.48550/arXiv.2509.21073},
  doi          = {10.48550/ARXIV.2509.21073},
  eprinttype    = {arXiv},
  eprint       = {2509.21073},
  bibsource    = {dblp computer science bibliography, https://dblp.org}
}

@article{tarasov2025nina,
  title={Nina: Normalizing flows in action. training vla models with normalizing flows},
  author={Tarasov, Denis and Nikulin, Alexander and Zisman, Ilya and Klepach, Albina and Lyubaykin, Nikita and Polubarov, Andrei and Derevyagin, Alexander and Kurenkov, Vladislav},
  journal={arXiv preprint arXiv:2508.16845},
  year={2025}
}

@article{kolesnikov2024jet,
  title={Jet: A modern transformer-based normalizing flow},
  author={Kolesnikov, Alexander and Pinto, Andr{\'e} Susano and Tschannen, Michael},
  journal={arXiv preprint arXiv:2412.15129},
  year={2024}
}

@article{dinh2016density,
  title={Density estimation using real nvp},
  author={Dinh, Laurent and Sohl-Dickstein, Jascha and Bengio, Samy},
  journal={arXiv preprint arXiv:1605.08803},
  year={2016}
}

@inproceedings{khader2021learning,
  title={Learning stable normalizing-flow control for robotic manipulation},
  author={Khader, Shahbaz Abdul and Yin, Hang and Falco, Pietro and Kragic, Danica},
  booktitle={2021 IEEE International Conference on Robotics and Automation (ICRA)},
  pages={1644--1650},
  year={2021},
  organization={IEEE}
}

@article{akimov2022let,
  title={Let offline rl flow: Training conservative agents in the latent space of normalizing flows},
  author={Akimov, Dmitriy and Kurenkov, Vladislav and Nikulin, Alexander and Tarasov, Denis and Kolesnikov, Sergey},
  journal={arXiv preprint arXiv:2211.11096},
  year={2022}
}

@inproceedings{mazoure2020leveraging,
  title={Leveraging exploration in off-policy algorithms via normalizing flows},
  author={Mazoure, Bogdan and Doan, Thang and Durand, Audrey and Pineau, Joelle and Hjelm, R Devon},
  booktitle={Conference on Robot Learning},
  pages={430--444},
  year={2020},
  organization={PMLR}
}

@article{chao2024maximum,
  title={Maximum entropy reinforcement learning via energy-based normalizing flow},
  author={Chao, Chen-Hao and Feng, Chien and Sun, Wei-Fang and Lee, Cheng-Kuang and See, Simon and Lee, Chun-Yi},
  journal={Advances in Neural Information Processing Systems},
  volume={37},
  pages={56136--56165},
  year={2024}
}

@article{black2024pi_0,
  title={$\pi_0$: A Vision-Language-Action Flow Model for General Robot Control},
  author={Black, Kevin and Brown, Noah and Driess, Danny and Esmail, Adnan and Equi, Michael and Finn, Chelsea and Fusai, Niccolo and Groom, Lachy and Hausman, Karol and Ichter, Brian and others},
  journal={arXiv preprint arXiv:2410.24164},
  year={2024}
}

@misc{christoph2025orcaopensourcereliablecosteffective,
      title={ORCA: An Open-Source, Reliable, Cost-Effective, Anthropomorphic Robotic Hand for Uninterrupted Dexterous Task Learning}, 
      author={Clemens C. Christoph and Maximilian Eberlein and Filippos Katsimalis and Arturo Roberti and Aristotelis Sympetheros and Michel R. Vogt and Davide Liconti and Chenyu Yang and Barnabas Gavin Cangan and Ronan J. Hinchet and Robert K. Katzschmann},
      year={2025},
      eprint={2504.04259},
      archivePrefix={arXiv},
      primaryClass={cs.RO},
      url={https://arxiv.org/abs/2504.04259}, 
}

@misc{nvidia2025isaaclabgpuacceleratedsimulation,
      title={Isaac Lab: A GPU-Accelerated Simulation Framework for Multi-Modal Robot Learning}, 
      author={Mayank Mittal and Pascal Roth and James Tigue and Antoine Richard and Octi Zhang and Peter Du and Antonio Serrano-Muñoz and Xinjie Yao and René Zurbrügg and Nikita Rudin and Lukasz Wawrzyniak and Milad Rakhsha and Alain Denzler and Eric Heiden and Ales Borovicka and Ossama Ahmed and Iretiayo Akinola and Abrar Anwar and Mark T. Carlson and Ji Yuan Feng and Animesh Garg and Renato Gasoto and Lionel Gulich and Yijie Guo and M. Gussert and Alex Hansen and Mihir Kulkarni and Chenran Li and Wei Liu and Viktor Makoviychuk and Grzegorz Malczyk and Hammad Mazhar and Masoud Moghani and Adithyavairavan Murali and Michael Noseworthy and Alexander Poddubny and Nathan Ratliff and Welf Rehberg and Clemens Schwarke and Ritvik Singh and James Latham Smith and Bingjie Tang and Ruchik Thaker and Matthew Trepte and Karl Van Wyk and Fangzhou Yu and Alex Millane and Vikram Ramasamy and Remo Steiner and Sangeeta Subramanian and Clemens Volk and CY Chen and Neel Jawale and Ashwin Varghese Kuruttukulam and Michael A. Lin and Ajay Mandlekar and Karsten Patzwaldt and John Welsh and Huihua Zhao and Fatima Anes and Jean-Francois Lafleche and Nicolas Moënne-Loccoz and Soowan Park and Rob Stepinski and Dirk Van Gelder and Chris Amevor and Jan Carius and Jumyung Chang and Anka He Chen and Pablo de Heras Ciechomski and Gilles Daviet and Mohammad Mohajerani and Julia von Muralt and Viktor Reutskyy and Michael Sauter and Simon Schirm and Eric L. Shi and Pierre Terdiman and Kenny Vilella and Tobias Widmer and Gordon Yeoman and Tiffany Chen and Sergey Grizan and Cathy Li and Lotus Li and Connor Smith and Rafael Wiltz and Kostas Alexis and Yan Chang and David Chu and Linxi "Jim" Fan and Farbod Farshidian and Ankur Handa and Spencer Huang and Marco Hutter and Yashraj Narang and Soha Pouya and Shiwei Sheng and Yuke Zhu and Miles Macklin and Adam Moravanszky and Philipp Reist and Yunrong Guo and David Hoeller and Gavriel State},
      year={2025},
      eprint={2511.04831},
      archivePrefix={arXiv},
      primaryClass={cs.RO},
      url={https://arxiv.org/abs/2511.04831}, 
}

@inproceedings{Luo2024SERL,
  author       = {Jianlan Luo and
                  Zheyuan Hu and
                  Charles Xu and
                  You Liang Tan and
                  Jacob Berg and
                  Archit Sharma and
                  Stefan Schaal and
                  Chelsea Finn and
                  Abhishek Gupta and
                  Sergey Levine},
  title        = {{SERL:} {A} Software Suite for Sample-Efficient Robotic Reinforcement
                  Learning},
  booktitle    = {{IEEE} International Conference on Robotics and Automation, {ICRA}
                  2024, Yokohama, Japan, May 13-17, 2024},
  pages        = {16961--16969},
  publisher    = {{IEEE}},
  year         = {2024},
  url          = {https://doi.org/10.1109/ICRA57147.2024.10610040},
  doi          = {10.1109/ICRA57147.2024.10610040},
  bibsource    = {dblp computer science bibliography, https://dblp.org}
}

@inproceedings{park2025flowQL,
  author       = {Seohong Park and
                  Qiyang Li and
                  Sergey Levine},
  title        = {Flow Q-Learning},
  booktitle    = {Forty-second International Conference on Machine Learning, {ICML}
                  2025, Vancouver, BC, Canada, July 13-19, 2025},
  publisher    = {OpenReview.net},
  year         = {2025},
  url          = {https://openreview.net/forum?id=KVf2SFL1pi},
  bibsource    = {dblp computer science bibliography, https://dblp.org}
}

@article{zhang2025reinflow,
  title={ReinFlow: Fine-tuning flow matching policy with online reinforcement learning},
  author={Zhang, Tonghe and Yu, Chao and Su, Sichang and Wang, Yu},
  journal={arXiv preprint arXiv:2505.22094},
  year={2025}
}

@article{black2025real,
  title={Real-Time Execution of Action Chunking Flow Policies},
  author={Black, Kevin and Galliker, Manuel Y and Levine, Sergey},
  journal={arXiv preprint arXiv:2506.07339},
}

@article{farebrother2024stop,
  title={Stop regressing: Training value functions via classification for scalable deep rl},
  author={Farebrother, Jesse and Orbay, Jordi and Vuong, Quan and Ta{\"\i}ga, Adrien Ali and Chebotar, Yevgen and Xiao, Ted and Irpan, Alex and Levine, Sergey and Castro, Pablo Samuel and Faust, Aleksandra and others},
  journal={arXiv preprint arXiv:2403.03950},
  year={2024}
}

@article{tarasov2024value,
  title={Is Value Functions Estimation with Classification Plug-and-play for Offline Reinforcement Learning?},
  author={Tarasov, Denis and Brilliantov, Kirill and Kharlapenko, Dmitrii},
  journal={arXiv preprint arXiv:2406.06309},
  year={2024}
}

@inproceedings{imani2018improving,
  title={Improving regression performance with distributional losses},
  author={Imani, Ehsan and White, Martha},
  booktitle={International conference on machine learning},
  pages={2157--2166},
  year={2018},
  organization={PMLR}
}

@article{tarasov2023corl,
  title={CORL: Research-oriented deep offline reinforcement learning library},
  author={Tarasov, Denis and Nikulin, Alexander and Akimov, Dmitry and Kurenkov, Vladislav and Kolesnikov, Sergey},
  journal={Advances in Neural Information Processing Systems},
  volume={36},
  pages={30997--31020},
  year={2023}
}

@article{levine2020offline,
  title={Offline reinforcement learning: Tutorial, review, and perspectives on open problems},
  author={Levine, Sergey and Kumar, Aviral and Tucker, George and Fu, Justin},
  journal={arXiv preprint arXiv:2005.01643},
  year={2020}
}

@article{tarasov2023revisiting,
  title={Revisiting the minimalist approach to offline reinforcement learning},
  author={Tarasov, Denis and Kurenkov, Vladislav and Nikulin, Alexander and Kolesnikov, Sergey},
  journal={Advances in Neural Information Processing Systems},
  volume={36},
  pages={11592--11620},
  year={2023}
}

@article{fujimoto2021minimalist,
  title={A minimalist approach to offline reinforcement learning},
  author={Fujimoto, Scott and Gu, Shixiang Shane},
  journal={Advances in neural information processing systems},
  volume={34},
  pages={20132--20145},
  year={2021}
}

@article{beeson2022improving,
  title={Improving td3-bc: Relaxed policy constraint for offline learning and stable online fine-tuning},
  author={Beeson, Alex and Montana, Giovanni},
  journal={arXiv preprint arXiv:2211.11802},
  year={2022}
}

@article{hu2023imitation,
  title={Imitation bootstrapped reinforcement learning},
  author={Hu, Hengyuan and Mirchandani, Suvir and Sadigh, Dorsa},
  journal={arXiv preprint arXiv:2311.02198},
  year={2023}
}

@inproceedings{ball2023efficient,
  title={Efficient online reinforcement learning with offline data},
  author={Ball, Philip J and Smith, Laura and Kostrikov, Ilya and Levine, Sergey},
  booktitle={International Conference on Machine Learning},
  pages={1577--1594},
  year={2023},
  organization={PMLR}
}

@article{nair2020awac,
  title={Awac: Accelerating online reinforcement learning with offline datasets},
  author={Nair, Ashvin and Gupta, Abhishek and Dalal, Murtaza and Levine, Sergey},
  journal={arXiv preprint arXiv:2006.09359},
  year={2020}
}

@article{nakamoto2023cal,
  title={Cal-ql: Calibrated offline rl pre-training for efficient online fine-tuning},
  author={Nakamoto, Mitsuhiko and Zhai, Simon and Singh, Anikait and Sobol Mark, Max and Ma, Yi and Finn, Chelsea and Kumar, Aviral and Levine, Sergey},
  journal={Advances in Neural Information Processing Systems},
  volume={36},
  pages={62244--62269},
  year={2023}
}

@article{smith2022walk,
  title={A walk in the park: Learning to walk in 20 minutes with model-free reinforcement learning},
  author={Smith, Laura and Kostrikov, Ilya and Levine, Sergey},
  journal={arXiv preprint arXiv:2208.07860},
  year={2022}
}

@article{oquab2023dinov2,
  title={Dinov2: Learning robust visual features without supervision},
  author={Oquab, Maxime and Darcet, Timoth{\'e}e and Moutakanni, Th{\'e}o and Vo, Huy and Szafraniec, Marc and Khalidov, Vasil and Fernandez, Pierre and Haziza, Daniel and Massa, Francisco and El-Nouby, Alaaeldin and others},
  journal={arXiv preprint arXiv:2304.07193},
  year={2023}
}

@article{loshchilov2017decoupled,
  title={Decoupled weight decay regularization},
  author={Loshchilov, Ilya and Hutter, Frank},
  journal={arXiv preprint arXiv:1711.05101},
  year={2017}
}

@inproceedings{handa2023dextreme,
  title={Dextreme: Transfer of agile in-hand manipulation from simulation to reality},
  author={Handa, Ankur and Allshire, Arthur and Makoviychuk, Viktor and Petrenko, Aleksei and Singh, Ritvik and Liu, Jingzhou and Makoviichuk, Denys and Van Wyk, Karl and Zhurkevich, Alexander and Sundaralingam, Balakumar and others},
  booktitle={2023 IEEE International Conference on Robotics and Automation (ICRA)},
  pages={5977--5984},
  year={2023},
  organization={IEEE}
}

@article{mandlekar2021matters,
  title={What matters in learning from offline human demonstrations for robot manipulation},
  author={Mandlekar, Ajay and Xu, Danfei and Wong, Josiah and Nasiriany, Soroush and Wang, Chen and Kulkarni, Rohun and Fei-Fei, Li and Savarese, Silvio and Zhu, Yuke and Mart{\'\i}n-Mart{\'\i}n, Roberto},
  journal={arXiv preprint arXiv:2108.03298},
  year={2021}
}

@misc{he2015deep,
  title={Deep residual learning for image recognition. CoRR abs/1512.03385 (2015)},
  author={He, Kaiming and Zhang, Xiangyu and Ren, Shaoqing and Sun, Jian},
  year={2015}
}

@article{tarasov2024role,
  title={The role of deep learning regularizations on actors in offline rl},
  author={Tarasov, Denis and Surina, Anja and Gulcehre, Caglar},
  journal={arXiv preprint arXiv:2409.07606},
  year={2024}
}

@article{park2024ogbench,
  title={Ogbench: Benchmarking offline goal-conditioned rl},
  author={Park, Seohong and Frans, Kevin and Eysenbach, Benjamin and Levine, Sergey},
  journal={arXiv preprint arXiv:2410.20092},
  year={2024}
}

@article{fu2020d4rl,
  title={D4rl: Datasets for deep data-driven reinforcement learning},
  author={Fu, Justin and Kumar, Aviral and Nachum, Ofir and Tucker, George and Levine, Sergey},
  journal={arXiv preprint arXiv:2004.07219},
  year={2020}
}

@article{intelligence2025pi,
  title={$\pi^{*}_{0.6}$: a VLA That Learns From Experience},
  author={Intelligence, Physical and Amin, Ali and Aniceto, Raichelle and Balakrishna, Ashwin and Black, Kevin and Conley, Ken and Connors, Grace and Darpinian, James and Dhabalia, Karan and DiCarlo, Jared and others},
  journal={arXiv preprint arXiv:2511.14759},
  year={2025}
}

@article{zhao2023learning,
  title={Learning fine-grained bimanual manipulation with low-cost hardware},
  author={Zhao, Tony and Kumar, Vikash and Levine, Sergey and Finn, Chelsea},
  journal={arXiv preprint arXiv:2304.13705},
  year={2023}
}

@article{zhou2017places,
  title={Places: A 10 million Image Database for Scene Recognition},
  author={Zhou, Bolei and Lapedriza, Agata and Khosla, Aditya and Oliva, Aude and Torralba, Antonio},
  journal={IEEE Transactions on Pattern Analysis and Machine Intelligence},
  year={2017},
  publisher={IEEE}
}

@article{sivakumar2022retargeter,
  title={Robotic telekinesis: Learning a robotic hand imitator by watching humans on youtube},
  author={Sivakumar, Aravind and Shaw, Kenneth and Pathak, Deepak},
  journal={arXiv preprint arXiv:2202.10448},
  year={2022}
}

@article{schwarke2025rslrl,
  title={Rsl-rl: A learning library for robotics research},
  author={Schwarke, Clemens and Mittal, Mayank and Rudin, Nikita and Hoeller, David and Hutter, Marco},
  journal={arXiv preprint arXiv:2509.10771},
  year={2025}
}

@article{miki2022learning,
  title={Learning robust perceptive locomotion for quadrupedal robots in the wild},
  author={Miki, Takahiro and Lee, Joonho and Hwangbo, Jemin and Wellhausen, Lorenz and Koltun, Vladlen and Hutter, Marco},
  journal={Science robotics},
  volume={7},
  number={62},
  pages={eabk2822},
  year={2022},
  publisher={American Association for the Advancement of Science}
}

@article{zhao2023ACT,
  title={Learning fine-grained bimanual manipulation with low-cost hardware},
  author={Zhao, Tony Z and Kumar, Vikash and Levine, Sergey and Finn, Chelsea},
  journal={arXiv preprint arXiv:2304.13705},
  year={2023}
}

@inproceedings{dasari2025ingredients,
  title={The ingredients for robotic diffusion transformers},
  author={Dasari, Sudeep and Mees, Oier and Zhao, Sebastian and Srirama, Mohan Kumar and Levine, Sergey},
  booktitle={2025 IEEE International Conference on Robotics and Automation (ICRA)},
  pages={15617--15625},
  year={2025},
  organization={IEEE}
}

@article{black2025training,
  title={Training-time action conditioning for efficient real-time chunking},
  author={Black, Kevin and Ren, Allen Z and Equi, Michael and Levine, Sergey},
  journal={arXiv preprint arXiv:2512.05964},
  year={2025}
}
